%% file: main_camera_ready.tex
\documentclass{article}

\usepackage{microtype}
\usepackage{graphicx}
\usepackage{subfigure}
\usepackage{booktabs} 

\usepackage{hyperref}

\usepackage[accepted]{icml2020}

\icmltitlerunning{Polynomial Tensor Sketch}

\usepackage{amssymb}
\usepackage{amsmath}
\usepackage{multicol}
\usepackage{multirow}
\usepackage{caption}
\usepackage{stackengine}
\usepackage{mathtools}
\usepackage{array}
\usepackage{setspace}
\numberwithin{equation}{section}
\input{notation.tex}

\newcolumntype{C}[1]{>{\centering\arraybackslash}p{#1}}
\begin{document}

\twocolumn[
\icmltitle{Polynomial Tensor Sketch  
for Element-wise Function of Low-Rank Matrix}



\icmlsetsymbol{equal}{*}

\begin{icmlauthorlist}
\icmlauthor{Insu Han}{kaist}
\icmlauthor{Haim Avron}{tau}
\icmlauthor{Jinwoo Shin}{kaistgai,kaist}
\end{icmlauthorlist}

\icmlaffiliation{kaist}{School of Electrical Engineering, KAIST, Daejeon, Korea}
\icmlaffiliation{tau}{School of Mathematical Sciences, Tel Aviv University, Israel}
\icmlaffiliation{kaistgai}{Graduate School of AI, KAIST, Daejeon, Korea}
\icmlcorrespondingauthor{Jinwoo Shin}{jinwoos@kaist.ac.kr}

\icmlkeywords{Machine Learning, ICML}

\vskip 0.3in
]



\printAffiliationsAndNotice{}  

\begin{abstract}
This paper studies how to sketch element-wise functions of low-rank matrices. 
Formally, given low-rank matrix $A=[A_{ij}]$ and scalar non-linear function 
$f$, we aim for finding  an approximated low-rank representation of the (possibly high-rank) 
matrix $[f(A_{ij})]$. To this end, we propose an efficient sketching-based algorithm  
whose complexity is significantly lower than the number of entries of $A$, 
i.e., it runs without accessing all entries of $[f(A_{ij})]$ explicitly.
The main idea underlying our method is to combine a polynomial approximation of $f$ with the existing 
tensor sketch scheme for approximating monomials of entries of $A$. 
To balance the errors of the two approximation components in an optimal manner,
we propose a novel regression formula to find polynomial coefficients given $A$ 
and $f$. In particular, we utilize a coreset-based regression 
with a rigorous approximation guarantee.
Finally, 
we demonstrate the applicability and superiority of the proposed scheme under 
various machine learning tasks.
\end{abstract}

\section{Introduction}
\label{sec:intro}
\input{intro.tex}

\section{Preliminaries}
\label{sec:prelim}
\input{prelim.tex}

\section{Linear-time Approximation of Element-wise Matrix Functions}
\label{sec:mainresult}
\input{mainresults.tex}

\input{coefficient.tex}

\section{Experiments}
\label{sec:app}
\input{exp.tex}

\section{Conclusion}
\label{sec:conclusion}
\input{conclusion.tex}

\section*{Acknowledgements}
IH and JS were partially supported by Institute of Information \& Communications Technology Planning \& Evaluation (IITP) grant funded by the Korea government (MSIT) (No.2019-0-00075, Artificial Intelligence Graduate School Program (KAIST)) and the Engineering Research Center Program through the National Research Foundation of Korea (NRF) funded by the Korean Government MSIT (NRF-2018R1A5A1059921).  HA was partially supported by BSF grant 2017698 and ISF grant 1272/17.

\bibliography{biblist}
\bibliographystyle{icml2020}

\clearpage
\appendix
\onecolumn
\icmltitle{Supplementary Material: \\
Polynomial Tensor Sketch for Element-wise Function of Low-Rank Matrix
}
\label{sec:extra_exp}
\input{extra_exp.tex}

\clearpage

\section{Proofs}
\label{sec:proofs}
\subsection{Proof of Proposition \ref{prop:pts}}
\input{proof1.tex}

\subsection{Proof of Lemma \ref{lmm:upperbound}}
\input{proof2.tex}

\subsection{Proof of Theorem \ref{thm:main}}
\input{proof3.tex}

\subsection{Proof of Theorem \ref{thm:coreset}}
\input{proof4.tex}

\end{document}

%% file: notation.tex
\newtheorem{theorem}{Theorem}
\newtheorem{proposition}[theorem]{Proposition}
\newtheorem{definition}[theorem]{Definition}
\newtheorem{lemma}[theorem]{Lemma}

\newenvironment{proof}{\setlength\parindent{0pt}{\bf Proof.    }}{\hfill\rule{2mm}{2mm}}

\newcommand{\R}{\mathbb{R}}

\renewcommand{\b}{\boldsymbol{b}}
\renewcommand{\c}{\boldsymbol{c}}

\newcommand{\f}{\boldsymbol{f}}

\renewcommand{\v}{\boldsymbol{v}}
\renewcommand{\u}{\boldsymbol{u}}

\newcommand{\x}{\boldsymbol{x}}
\newcommand{\y}{\boldsymbol{y}}

\renewcommand{\a}{\boldsymbol{a}}

\newcommand{\bu}{\overline{\u}}
\newcommand{\bv}{\overline{\v}}

\newcommand{\abs}[1]{\left|{#1}\right|}
\newcommand{\norm}[1]{\left\lVert#1\right\rVert}

\newcommand{\inner}[1]{\left \langle {#1} \right \rangle}
\newcommand{\fele}[1]{f^{\odot}({#1})}

\DeclareMathOperator*{\argmin}{arg\,min}

\newcommand{\bigo}[1]{\mathcal{O}(#1)}
\newcommand{\CS}{{\sc CountSketch} }
\newcommand{\TS}{{\sc TensorSketch }}
\newcommand{\PTS}{{\sc Poly-TensorSketch }}
\newcommand{\FFT}[1]{{\textsc{FFT}}(#1)}
\newcommand{\IFFT}[1]{{\textsc{FFT}}^{-1}\left(#1\right)}

\newcommand{\segment}{~$\mathtt{segment}$~}
\newcommand{\satimage}{~$\mathtt{satimage}$~}
\newcommand{\usps}{~$\mathtt{usps}$~}
\newcommand{\anuran}{~$\mathtt{anuran}$~}
\newcommand{\grid}{~$\mathtt{grid}$~}
\newcommand{\mapping}{~$\mathtt{mapping}$~}

\newcommand{\letter}{~$\mathtt{letter}$~}

%% file: intro.tex
Given a low-rank matrix $A = [A_{ij}]\in \R^{n \times n}$ 
with $A=UV^\top$ for some matrices $U,V\in \R^{n\times d}$ with $d\ll n$
and a scalar non-linear function $f: \mathbb{R} \rightarrow \mathbb{R}$, 
we are interested in the following element-wise matrix function:\footnotemark
\begin{align*}
\fele{A} \in \R^{n \times n},\quad \text{where } \ \fele{A}_{ij} := 
\left[f(A_{ij})\right].
\end{align*}
Our goal is to design a fast algorithm for computing tall-and-thin matrices $T_U, 
T_V$ in time $o(n^2)$
such that $\fele{A}\approx T_UT_V^\top$. In particular, our algorithm  
should run without computing all entries of $A$ or 
$\fele{A}$ explicitly. As an side, we obtain a $o(n^2)$-time approximation scheme of
$\fele{A} \x \approx T_UT_V^\top \x$ for an arbitrary vector  
$\x \in \R^n$ due to the associative property of matrix 
multiplication, where the exact computation of $\fele{A} \x$ requires the 
complexity of $\Theta(n^2)$.
\footnotetext{In this paper, we primarily focus on the square matrix $A$ for simplicity, but it is straightforward to extend our results to the case of non-square matrices.}


The matrix-vector multiplication $\fele{A} \x$ or the low-rank decomposition 
$\fele{A}\approx T_UT_V^\top$ is useful in many machine learning algorithms. 
For example, the Gram matrices of certain kernel functions, e.g., polynomial 
and radial basis function (RBF), are element-wise matrix functions where the 
rank is the dimension of the underlying data. Such matrices are the cornerstone of 
so-called kernel methods and the ability to multiply the Gram matrix to a 
vector suffices for most kernel learning. The Sinkhorn-Knopp algorithm is a 
powerful, yet simple, tool for computing optimal transport distances 
\cite{cuturi2013sinkhorn, altschuler2017near} and also involves the 
matrix-vector multiplication $\fele{A} \x$ with $f(x)=\exp(x)$. 
Finally, $\fele{UV^\top}$ can also describe the 
non-linear computation of activation in a layer of deep neural networks 
\cite{goodfellow2016deep}, where $U$, $V$ and $f$ correspond to input, weight 
and activation function (e.g., sigmoid or ReLU) of the previous layer, 
respectively. 

{
Unlike the element-wise matrix function $\fele{A}$, 
a traditional matrix function $f(A)$ is defined on their eigenvalues \cite{higham2008functions} and possesses clean algebraic properties, i.e., it preserves eigenvectors.
For example, given a diagonalizable matrix $A = PDP^{-1}$, it is defined that $f(A) = P f^{\odot}(D) P^{-1}$. 
A classical problem addressed in the literature is of approximating the trace of $f(A)$
(or $f(A) \x$)   
efficiently \cite{han2016approximating, ubaru2017fast}. 
However, these methods are not applicable to our problem 
because element-wise matrix functions are fundamentally different from the traditional function of matrices, 
e.g., they do not guarantee the spectral property.
We are unaware of any approximation algorithm that targets general element-wise 
matrix functions. 
The special case of element-wise kernel functions such as
$f(x)=\exp(x)$ and $f(x)=x^k$, as been address in the literature, e.g., using
random Fourier features ({\sc RFF}) 
\cite{rahimi2008random, pennington2015spherical},
{
Nystr\"{o}m method \cite{williams2001using}, sparse greedy approximation 
\cite{smola2000sparse} and sketch-based methods \cite{pham2013fast, 
avron2014subspace, meister2019tight, ahle2020oblivious}}, just to name a few examples.
We aim for not only designing an 
approximation framework for general $f$, but also outperforming the previous 
approximations (e.g., {\sc RFF}) even for the special case $f(x)=\exp(x)$. 
}

The high-level idea underlying our method is to combine two approximation schemes:
\TS approximating the element-wise matrix function of monomial $x^k$ 
and a degree-$r$ polynomial approximation $p_r(x)=\sum_{j=0}^r c_j x^j$ of 
function $f$, e.g., $f(x) = \exp(x) \approx 
p_r(x)=\sum_{j=0}^r\frac{1}{j!}x^j$. 
More formally, we consider the following approximation:
\begin{align*}
\fele{A} 
&~\overset{\text{(a)}}{\approx}~ \sum_{j=0}^r c_j \cdot (x^j)^{\odot}(A) \\ 
&~\overset{\text{(b)}}{\approx}~ \sum_{j=0}^r c_j \cdot 
\text{\TS}\left((x^j)^{\odot}(A)\right),
\end{align*}
which we call {\sc Poly-TensorSketch}.
This is a linear-time approximation scheme with respect to $n$ under the choice of $r=\mathcal{O}(1)$. 
Here, a non-trivial challenge occurs:
a larger degree $r$ is required to approximate an arbitrary function $f$ better,
while the approximation error of {\sc TensorSketch} is known to increase exponentially with respect to $r$ \cite{pham2013fast, avron2014subspace}.
Hence, it is important
to choose good $c_j$'s for balancing two approximation errors in both (a) and (b).
The known (truncated) polynomial expansions such as Taylor or Chebyshev \cite{mason2002chebyshev} are far from being optimal for this purpose (see
Section \ref{sec:polyreg} and \ref{sec:exp1} for more details). 

{
To address this challenge, we formulate an optimization problem on the polynomial coefficients (the $c_j$s)
by relating them to the approximation error of {\sc Poly-TensorSketch}.
However, this optimization problem is intractable since the objective 
involves an expectation taken over random variables whose supports are exponentially large. 
Thus, we derive a novel tractable alternative optimization problem based on an upper bound the approximation error of {\sc Poly-TensorSketch}.
This problem turns out to be an instance of generalized ridge regression \cite{hemmerle1975explicit},
and as such has a closed-form solution. 
Furthermore, we observe that the regularization term effectively forces the coefficients to be exponentially decaying, and this compensates the exponentially growing errors of {\sc TensorSketch} with respect to the rank, while simultaneously maintaining a good polynomial approximation to the scalar function $f$ given entries of $A$.
We further reduce its complexity by regressing only a subset of the matrix entries
found by 
an efficient coreset clustering algorithm \cite{har2008geometric}: 
if matrix entries are close to the selected coreset, 
then the resulting coefficient is provably close to the optimal one.
Finally, we construct the regression with respect to Chebyshev polynomial basis (instead of monomials), i.e.,
$p_r(x) = \sum_{j=0}^r {c}_j t_j(x)$
for the Chebyshev polynomial $t_j(x)$ of degree $j$,
in order to resolve numerical issues arising for large degrees. 
}

We evaluate the approximation quality of
our algorithm under the {\sc RBF} kernels of synthetic and real-world datasets. Then, 
we apply the proposed method to 
classification using kernel {\sc SVM} \cite{cristianini2000introduction,scholkopf2001learning}
and computation of optimal transport distances \cite{cuturi2013sinkhorn} 
which require to compute element-wise matrix functions with $f(x)=\exp(x)$.
Our experimental results confirm that our scheme is at the order of magnitude faster than the exact method 
with a marginal loss on accuracy. Furthermore, our scheme also significantly outperforms a state-of-the-art approximation method, {\sc RFF} for the aforementioned applications.
Finally, we demonstrate a wide applicability of our method by applying it to the linearization of neural networks, which 
requires to compute element-wise matrix functions with $f(x)=\mathtt{sigmoid}(x)$.


%% file: prelim.tex
In this section, we provide backgrounds on the randomized sketching algorithms \CS and \TS that
are crucial components of the proposed scheme.

First, \CS \cite{charikar2002finding, weinberger2009feature} 
was proposed for an effective dimensionality reduction of high-dimensional 
vector $\u \in \mathbb{R}^{d}$. Formally, consider 
a random hash function $h : [d] \rightarrow [m]$ and a random sign function $s: [d] \rightarrow \{-1,+1 \}$, where $[d]:=\{1,\dots, d\}$.
Then, \CS transforms $\u$ into $C_{\u} \in \R^m$ such that 
$[C_{\u}]_j := \sum_{i: h(i) = j} s(i) {u}_i$
for $j \in [m]$.
The algorithm takes $\mathcal{O}(d)$ time to run since it requires a single pass over the input. 
It is known that applying the same\footnote{i.e., use the same hash and sign functions.} \CS transform on two vectors preserves the dot-product, i.e., 
$\inner{\u, \v} = \mathbf{E}\left[ \inner{C_{\u}, C_{\v}}\right]$.


\TS \cite{pham2013fast} was proposed as a generalization of \CS to tensor products of vectors.
Given $\u \in \R^d$ and a degree $k$, consider i.i.d. random hash functions 
$h_1, \dots, h_k : [d] \rightarrow [m]$ and i.i.d. random sign functions $s_1, 
\dots, s_k:[d]\rightarrow \{-1,+1\}$, \TS applied to $\u$ is defined as the $m$-dimensional vector
$T_{\u}\in \R^m$ such that $j\in [m]$,
\begin{align*}
[ T_{\u}^{(k)} ]_j := \sum_{H(i_1, \dots, i_k) = j} s_1(i_1) \dots s_k(i_k) 
{u}_{i_1} \dots {u}_{i_k},
\end{align*}
where
$H(i_1, \dots, i_k) \equiv h_1(i_1) + \dots + h_k(i_k) \ \ \text{(mod } \ m).$\footnotemark
\footnotetext{Unless stated otherwise, we define $T_{\u}^{(0)} := 1$.}
In \cite{pham2013fast, avron2014subspace, pennington2015spherical, 
meister2019tight}, \TS was 
used for approximating of the power of dot-product between vectors.  In other 
words, let $T_{\u}^{(k)}, T_{\v}^{(k)}$ be the same \TS on $\u, \v \in 
\R^d$ with degree $k$. Then, it holds that 
\begin{align*}
\inner{\u, \v}^k
= 
\inner{\u^{\otimes k}, \v^{\otimes k}}
=
\mathbf{E}\left[\inner{T_{\u}^{(k)}, T_{\v}^{(k)}}\right],
\end{align*}
where $\otimes$ denotes the tensor product (or outer-product) and $\u^{\otimes 
k} := \u \otimes \dots \otimes \u \in \R^{d^k}$ ($k-1$ times).  This can be 
naturally extended to matrices $U,V \in \R^{n \times d}$. 
Suppose $T_U^{(k)}, T_V^{(k)} \in \R^{n \times m}$ are the same \TS on each row 
of $U,V$, and it follows that 
$(UV^\top)^{\odot k} = \mathbf{E}\left[ T_U^{(k)} T_V^{(k) \top}\right]$
where $A^{\odot k} := [A^k_{ij}]$.
\citet{pham2013fast} devised a fast way to 
compute \TS using the Fast Fourier Transform ({\sc FFT}) as described in 
Algorithm \ref{alg:ts}.
\begin{algorithm}[t]
\caption{\TS \cite{pham2013fast}  }\label{alg:ts}
\begin{algorithmic}[1]
\STATE {\bf Input}: $U\in \mathbb{R}^{n \times d}$, degree $k$, sketch dimension $m$
\STATE Draw $k$ i.i.d. random hash functions $h_1, \dots, h_k : [d] \rightarrow 
[m]$ 
\STATE Draw $k$ i.i.d. random sign functions $s_1, \dots, s_k :[d]\rightarrow 
\{+1,-1\}$
\STATE $C_U^{(i)}$ $\leftarrow$ \CS on each row of $U$ using $h_i$ and $s_i$ for $i = 1, \dots ,k$.
\STATE $T_U^{(k)} \leftarrow  \text{\sc FFT}^{-1}\left(
\text{\sc FFT}(C_U^{(1)}) \odot \cdots \odot \text{\sc FFT}(C_U^{(k)}) 
\right)$
\STATE {\bf Output} : $T_U^{(k)}$
\end{algorithmic}
\end{algorithm}

In Algorithm \ref{alg:ts}, $\odot$ is the element-wise multiplication (also called
the Hadamard product) between two matrices of the same dimension 
and $\text{\sc FFT}(C), \text{\sc FFT}^{-1}(C)$ are the Fast Fourier Transform 
and its inverse applied to each row of a matrix $C\in \mathbb{R}^{n\times m}$, respectively.
The cost of the FFTs is $\mathcal O(n m\log m)$ and so the total cost of 
Algorithm \ref{alg:ts} is $\mathcal{O}(nk(d + m \log m) )$ time
since it runs {\sc FFT} and \CS $k$ times.
\citet{avron2014subspace} proved a tight bound on the variance (or 
error) of \TS as follows.

\begin{theorem}[\citet{avron2014subspace}] \label{thm:tensor}
Given $U, V \in \mathbb{R}^{n \times d}$, let $T_U^{(k)}, T_V^{(k)} \in \mathbb{R}^{n \times m}$ be the same {\sc TensorSketch} of $U,V$ with degree $k \geq 0$ and sketch dimension $m$. Then, it holds
\begin{align} \label{eq:tsbound}
&\mathbf{E} \left[ 
\left\| \left( UV^\top\right)^{\odot k}
- T_U^{(k)} {T_V^{(k)}}^\top
\right\|_F^2
\right] \nonumber\\
&\leq 
\frac{(2 + 3^k) \left( \sum_i (\sum_{j} U_{ij}^{2})^k \right) \left( \sum_i (\sum_{j} V_{ij}^{2})^k \right)}{m}.
\end{align}
\end{theorem}
The above theorem implies the error of \TS becomes small for large sketch 
dimension $m$, but can grow fast with respect to degree $k$. 
Recently, \citet{ahle2020oblivious} proposed another method whose error bound  
is tighter with respect to $k$, but can be worse with respect to another factor so-called statistical dimension \cite{avron2017sharper}.\footnotemark

%% file: mainresults.tex
Given a scalar function $f:\R\rightarrow \R$ 
and matrices $U, V \in \R^{n \times d}$ with $d\ll n$, 
our goal is to design an efficient algorithm 
to find $T_U, T_V\in \R^{n \times d^\prime}$
with $d^\prime \ll n$ in $o(n^2)$ time such that 
\begin{align*}
f^{\odot}(UV^\top) \approx T_U T_V^\top \in \R^{n \times n}.
\end{align*}
Namely, we aim for finding a low-rank approximation of $f^{\odot}(UV^\top)$
without computing all $n^2$ entries of $UV^\top$.
We first describe the proposed approximation scheme in Section \ref{sec:polyts}
and provide further ideas for minimizing the approximation gap in Section \ref{sec:polyreg}.
\footnotetext{In our experiments, \TS and the method by \citet{ahle2020oblivious}
show comparable approximation errors, but the latter one requires up to $4$ times more computation.}
\begin{algorithm}[t]
\caption{\PTS}\label{alg:polyts}
\begin{algorithmic}[1]
\STATE {\bf Input}: 
\ $U, V \in \R^{n \times d}$, degree $r$, coefficient $c_0, \dots, c_r$, 
sketch dimension $m$ 
\STATE Draw $r$ i.i.d. random hash functions $h_1, \dots, h_r : [d] 
\rightarrow [m]$ 
\STATE Draw $r$ i.i.d. random sign functions $s_1, \dots, s_r : 
[d]\rightarrow \{+1,-1\}$
\STATE $T_U^{(1)}, T_V^{(1)} \leftarrow $ \CS of $U,V$ using $h_1$ and $s_1$, respectively.
\STATE $F_U, F_V \leftarrow$ $\FFT{T_U^{(1)}}$, \ $\FFT{T_V^{(1)}}$
\STATE $\Gamma \leftarrow c_0 T_U^{(0)} T_V^{(0) \top} + c_1 T_U^{(1)} T_V^{(1) 
\top} $ 
\FOR{$j = 2$ to $r$}
\STATE $C_U, C_V \leftarrow $ \CS of $U,V$ using $h_j$ and $s_j$, respectively.
\STATE $F_U, F_V \leftarrow \FFT{C_U} \odot F_U, \ \FFT{C_V} \odot F_V$
\STATE $T_U^{(j)}, T_V^{(j)} \leftarrow \IFFT{F_U}, \ \IFFT{F_V}$
\STATE $\Gamma \leftarrow \Gamma + c_{j} T_U^{(j)} {T_V^{(j) \top}}$
\ENDFOR
\STATE {\bf Output} : $\Gamma$ 
\end{algorithmic}
\end{algorithm}

\subsection{\PTS Transform} \label{sec:polyts}
Suppose we have a polynomial $p_r(x) = \sum_{j=0}^r c_j x^j$ approximating 
$f(x)$, e.g., $f(x) = \exp(x)$ and its (truncated) Taylor series 
$p_r(x)=\sum_{j=0}^r\frac{1}{j!}x^j$. Then, we consider the following 
approximation scheme, coined \PTS:
\begin{align} \label{eq:approx1}
\fele{UV^\top} 
&~\overset{\text{(a)}}{\approx}~ \sum_{j=0}^r c_j (UV^\top)^{\odot j} 
~\overset{\text{(b)}}{\approx}~ \sum_{j=0}^r c_j T_U^{(j)} T_V^{(j)\top},
\end{align}
where $T_U^{(j)}, T_V^{(j)} \in \R^{n \times m}$ are the same \TS of $U, V$ 
with degree $j$ and
sketch dimension $m$, respectively. 
Namely, our main idea is to combine 
(a) a polynomial approximation of a scalar function with 
(b) the randomized tensor sketch of a matrix.
Instead of running Algorithm \ref{alg:ts} independently for each $j\in [r]$,
we utilize 
the following recursive relation to amortize operations: 
\begin{align*}
T_U^{(j)} = \IFFT{\FFT{C_U} \odot \FFT{T_U^{(j-1)}}},
\end{align*}
where $C_U$ is the \CS on each row of $U$ whose randomness is independently 
drawn from that of $T_U^{(j-1)}$.
Since each recursive step can be computed in $\mathcal{O}(n(d + m \log m))$ time, 
computing all $T_U^{(j)}$ for $j\in [r]$ requires
$\mathcal{O}(nr(d + m \log m))$ operations. 
Hence, the overall complexity of {\sc Poly-TensorSketch}, formally described in Algorithm \ref{alg:polyts}, is $\mathcal O(n)$ if $r,d,m=\mathcal O(1)$.
%

{
Observe that multiplication of $\Gamma$ (i.e., the output of Algorithm 
\ref{alg:polyts}) and an arbitrary vector $\x \in \R^d$
}
can be done in $\mathcal{O}(nmr)$ time due to
$\Gamma \x =\sum_{j=0}^r c_j T_U^{(j)} T_V^{(j) \top} \x$.
Hence, for $r,m,d =\mathcal O(1)$, 
{\sc Poly-TensorSketch} can approximate $f^{\odot}(UV^\top) \x$
in $\mathcal O(nr(d+m\log m))=\mathcal O(n)$ time.
As for the error, we prove the following error bound. 



\begin{proposition}\label{prop:pts}
Given $U, V \in \R^{n \times d}$, $f : \R\rightarrow \R$,
suppose 
that $| f(x) - \sum_{j=0}^r c_j x^j | \leq \varepsilon$ 
in a closed interval containing all entries of $UV^\top$ for some $\varepsilon >0$. 
Then, it holds that
\begin{align} \label{eq:errbound2}
&\mathbf{E}\left[\left\|
f^{\odot}(UV^\top)
- \Gamma 
\right\|_F^2
\right] 
\leq 2n^2 \varepsilon^2 \nonumber \\
&+ \sum_{j=1}^r  \frac{2 r c_j^2 (2+3^j) \left( \sum_i (\sum_{k} U_{ik}^{2})^j 
\right) \left( \sum_i (\sum_{k} V_{ik}^{2})^j \right)}{m},
\end{align}
where 
$\Gamma$ is the output of Algorithm \ref{alg:polyts}.
\end{proposition}


The proof of Proposition \ref{prop:pts} is given in the supplementary material.
Note that even when $\varepsilon$ is close to $0$,
the error bound \eqref{eq:errbound2} may increase exponentially with respect to the degree $r$.
This is because the approximation error of \TS grows exponentially with respect to the degree (see Theorem \ref{thm:tensor}). 
Therefore, in order to compensate, it is desirable to use exponentially decaying (or even truncated) coefficient $\{c_j\}$. However, we are now
faced with the challenge of finding exponentially decaying coefficients, while not hurting the approximation quality of $f$.
Namely, we want to balance the
two approximation components of
{\sc Poly-TensorSketch}: {\sc TensorSketch}
and a polynomial approximation for $f$.
In the following section, we propose a novel approach to find the optimal coefficients minimizing the approximation error 
of {\sc Poly-TensorSketch}.
\subsection{{Optimal Coefficient via Ridge Regression}} \label{sec:polyreg}



A natural choice for the coefficients is to utilize a polynomial series such as 
Taylor, Chebyshev \cite{mason2002chebyshev} or other orthogonal basis 
expansions (see \citet{szeg1939orthogonal}).
However, constructing the coefficient in this way focuses on the error of the 
polynomial approximation of $f$, and ignores the error of \TS which depends on 
the decay of the coefficient, as reflected in the error bound 
\eqref{eq:errbound2}. 
To address the issue, we study the following optimization to find the  optimal coefficient:
\begin{align}
\min_{\c \in \R^{r+1}} \mathbf{E}\left[
\norm{f^{\odot}(UV^\top) - \Gamma}_F^2
\right], \label{eq:idealopt}
\end{align}
where 
$\Gamma$ is the output of \PTS and $\c = [c_0, \dots, c_r]$ is a vector of the coefficient. 
However, it is not easy to solve the above optimization directly as its 
objective involves an expectation over random variables with a huge support, 
i.e., uniform hash and binary sign functions. Instead, we aim for minimizing 
an upper bound of the approximation error \eqref{eq:idealopt}.
To this end, we define the following notation.
\begin{definition}\label{def:not}
Let $X \in \R^{n^2 \times (r+1)}$ be the matrix\footnote{$X$ is known 
as the Vandermonde matrix.} whose $k$-th column corresponds to the  
vectorization of $(UV^\top)_{ij}^{k-1}$ for all $i,j \in [n]$, $\f \in 
\R^{n^2}$ be the vectorization of $\fele{UV^\top}$ and $W \in \R^{(r+1) 
\times (r+1)}$ be a diagonal matrix such that $W_{11} = 0$ and for $i=2, \dots, 
r+1$
\begin{align*} 
W_{ii} = \sqrt{\frac{r(2+3^i)(\sum_j (\sum_{k} U_{jk}^{2})^i)(\sum_j (\sum_{k} 
V_{jk}^{2})^i)}{m}}.
\end{align*} 
\end{definition}
Using the above notation, 
we establish the following error bound.

\begin{lemma} \label{lmm:upperbound}
Given $U, V \in \R^{n \times d}$ and
$f : \R\rightarrow \R$, 
consider $X, \f$ and $W$ defined in Definition \ref{def:not}.
Then, it holds 
\begin{align} \label{eq:errbound}
\mathbf{E}\left[\norm{f^{\odot}(UV^\top) - \Gamma}_F^2\right]
\leq  2 \|X\c - \f\|_2^2 + 2 \|W\c\|_2^2,
\end{align}
where $\Gamma$ is the output of Algorithm \ref{alg:polyts}.
\end{lemma}

{The proof of Lemma \ref{lmm:upperbound} is given in the supplementary material.}
Observe that the error bound \eqref{eq:errbound} is a quadratic form of $\c \in 
\R^{r+1}$, where it is straightforward to obtain a closed-form solution for 
minimizing \eqref{eq:errbound}: 
\begin{align} \label{eq:optc}
\c^* 
&:= \argmin_{\c \in \R^{r+1}}  \| X \c - \f \|_2^2 + \| W \c \|_2^2 \nonumber \\
&= \left( X^\top X + W^2\right)^{-1} X^\top \f.
\end{align}

This optimization task 
is also known as generalized ridge regression \cite{hemmerle1975explicit}.
The solution \eqref{eq:optc} minimizes the regression error (i.e., the error of polynomial), while
it is regularized by $W$, i.e., $W_{ii}$ is a regularizer of $c_{i}$.
Namely, if $W_{ii}$ grows exponentially with respect to $i$, 
then $c_{i}^*$ may decay exponentially (this compensates the error of \TS with degree $i$).
By substituting $\c^*$ 
into the error bound \eqref{eq:errbound}, 
we obtain the following multiplicative error bound of {\sc Poly-TensorSketch}.

%

\begin{theorem} \label{thm:main}
Given $U, V \in \R^{n \times d}$ and
$f : \R\rightarrow \R$, 
consider $X$ defined in Definition \ref{def:not}.
Then, it holds
\begin{align} \label{eq:errbound3}
\mathbf{E}\left[\norm{f^{\odot}(UV^\top) - \Gamma }_F^2\right]
\leq 
\left(\frac{2}{1 + \frac{m {\sigma}^2}{r C}}\right)
\|f^{\odot}(UV^\top) \|_F^2,
\end{align}
where 
$\Gamma$ is 
the output of Algorithm \ref{alg:polyts} with the coefficient $\c^*$ in 
\eqref{eq:optc},
$\sigma\geq 0$ is the smallest singular value of $X$ and 
$C = \max ( 5 \|U\|_F^2 \|V\|_F^2, (2+3^r)  (\sum_j (\sum_{k} U_{jk}^{2})^r)(\sum_j (\sum_{k} V_{jk}^{2})^r))$.
\end{theorem}

{The proof of Theorem \ref{thm:main} is given in the supplementary material.}
Observe that the error bound \eqref{eq:errbound3} is bounded by 
{2$\| f^{\odot}(UV^\top) \|_F^2$} since $m, r, \sigma, C\geq 0$.
On the other hand, 
we recall that the error bound \eqref{eq:errbound2}, i.e., {\sc 
Poly-TensorSketch} without using the optimal coefficient $\c^*$, 
can grow exponentially with respect to $r$.\footnote{
{Note that $\sigma$ is decreasing with respect to $r$, and
the error bound \eqref{eq:errbound3} may decrease with respect to $r$.
}
}
We indeed observe that 
$\c^*$ is empirically
superior to the coefficient of the popular Taylor and Chebyshev series expansions with respect to the
error of {\sc Poly-TensorSketch} 
(see Section \ref{sec:exp1} for more details).
We also remark that the error bound \eqref{eq:errbound3}
of {\sc Poly-TensorSketch} is even better than that 
\eqref{eq:tsbound} of {\sc TensorSketch} even for the case of monomial $f(x) = x^k$.
This is primarily because the former is achievable 
by paying an additional cost for computing  
the optimal coefficient vector \eqref{eq:optc}.
In what follows, we discuss the {extra} cost.

%% file: coefficient.tex
\subsection{{Reducing Complexity via Coreset Regression}} 
\label{sec:coreset}
To obtain the optimal coefficient $\c^*$ in \eqref{eq:optc}, 
one can check that $\mathcal{O}(r^2 n^2 + r^3)$ operations are required 
because of computing $X^\top X$
for $X \in \R^{n^2 \times (r+1)}$ (see Definition \ref{def:not}).
This hurts the overall complexity of 
{\sc Poly-TensorSketch}.
Instead, 
we choose a subset of entries in $UV^\top$ 
%
and approximately find the coefficient $\c^{*}$ 
based on the selected entries. 

\begin{algorithm}[t]
\caption{Greedy $k$-center} \label{alg:kcenter}
\begin{algorithmic}[1]
\STATE {\bf Input}: $\u_1, \dots, \u_n \in \R^d$, 
number of clusters $k$
\STATE $a \leftarrow$ uniformly random in $\{1, 2, \dots, n\}$ and $S \leftarrow \{a \}$ 
\STATE $\Delta_i \leftarrow \norm{\u_i - \u_a}_2$ for all $i\in [n]$
\FOR{$i = 2$ to $k$  }
\STATE $a \leftarrow \arg \max_{i} \Delta_i$ 
and $S \leftarrow S \cup \{ a\}$
\STATE $\Delta_i \leftarrow \min(\Delta_i, \norm{\u_i - \u_{a}}_{2} )$ for all $i$
\ENDFOR
\STATE $P(i) \leftarrow \arg\min_{j\in S} \norm{\u_i - \u_j}_{2}$ for all $i$
\STATE {\bf Output}: $\{ \u_i : i \in S\}, \ P$
\end{algorithmic}
\end{algorithm}
\begin{algorithm}[t]
\setstretch{1.2}
\caption{Coefficient approximation via coreset} \label{alg:coreset}
\begin{algorithmic}[1]
\STATE {\bf Input}: \ $U, V \in \R^{n \times d}$, a function $f$, number 
of clusters $k$, degree $r$
\STATE $\{\bu_i\}_{i=1}^k, P_1 \leftarrow $ 
Algorithm \ref{alg:kcenter} with rows in $U$ and $k$
\STATE $\{\bv_i\}_{i=1}^k, P_2 \leftarrow$ 
Algorithm \ref{alg:kcenter} with rows in $V$ and $k$
\STATE $\varepsilon_U \leftarrow \sum_i \norm{\u_i - \overline{\u}_{P_1(i)}}_2$, 
$\varepsilon_V \leftarrow \sum_i \norm{\v_i - \overline{\v}_{P_2(i)}}_2$
\IF{$\varepsilon_U \sum_i \norm{\v_i}_2 < \varepsilon_V \sum_j \norm{\u_j}_2$}
\STATE 
$\overline{X}, D$ and $\overline{\f} \leftarrow$ matrices and vector defined in Definition \ref{def:notcoreset} with $\overline{U}, V$ and $P_1$
\ELSE
\STATE 
$\overline{X}, D$ and $\overline{\f} \leftarrow$ matrices and vector defined in Definition \ref{def:notcoreset} with $U, \overline{V}$ and $P_2$
\ENDIF 
\STATE {\bf Output} : $\left(\overline{X}^\top D \overline{X} + 
W^2\right)^{-1} \overline{X}^\top D \overline{\f}$
\end{algorithmic}
\vspace{-0.03in}
\end{algorithm}

More formally, suppose $\overline{U} \in \R^{k \times d}$ is a matrix containing certain $k$ rows of $U  \in \R^{n \times d}$ 
and we use the following approximation:
\begin{align} \label{eq:coeffcoreset}
\c^* &=\left( X^\top X + W^2\right)^{-1} X^\top \f\notag\\
&\approx \left(\overline{X}^\top D \overline{X} + W^2\right)^{-1} \overline{X}^\top D \overline{\f},
\end{align}
where $\overline{X}, D$ and $\overline{\f}$ are defined as follows.\footnote{Note that $\overline{X}, \overline{\f}$ are defined similar to $X,f$ in Definition \ref{def:not}.}
{
\begin{definition} \label{def:notcoreset}
Let $\overline{X} \in \R^{kn \times (r+1)}$ be the matrix whose $\ell$-th column
is the vectorization of $(\overline{U}V^\top)_{ij}^{\ell-1}$ for $i \in [k], j \in [n]$ 
and $\overline{\f} \in \R^{kn}$ be the vectorization of $f^{\odot}(\overline{U}V^\top)$.
Given a mapping $P: [n]\rightarrow[k]$, 
let $D \in \R^{kn \times kn}$ be a diagonal matrix where $D_{ii} = \left|\{ s: P(s)=\lceil i/n \rceil\}\right|$.
\end{definition}
}
Computing \eqref{eq:coeffcoreset} requires $\bigo{r^2k n + r^3}=\bigo{n}$ 
for $r,k = \bigo{1}$.
In what follows, we show that if
rows of $U$ are close to those of $\overline{U}$ 
under the mapping $P$,
then the approximation \eqref{eq:coeffcoreset} becomes tighter.
To this end, we say $\overline{U}$ is a $\varepsilon$-coreset
of $U$ for some $\varepsilon>0$ 
if there exists a mapping $P : [n] \rightarrow [k]$ such that
$\sum_{i=1}^n \norm{\u_i - \overline{\u}_{P(i)}}_2 \leq \varepsilon,$
where $\u_i, \overline{\u}_{i}$ are the $i$-th rows of $U, \overline{U}$,
respectively. Using the notation,
we now present the following error bound of {\sc Poly-TensorSketch} under the approximated coefficient vector in \eqref{eq:coeffcoreset}.

\begin{theorem} \label{thm:coreset}
Given $U, V\in \mathbb{R}^{n \times d}$ and
$f : \R\rightarrow \R$,
consider $\varepsilon$-coreset $\overline{U}\in \mathbb{R}^{k \times d}$ of $U$
with $\overline{X}, D,\overline{\f}$ and $P$ defined in
Definition \ref{def:notcoreset}.
For the approximated coefficient vector $\c$ in \eqref{eq:coeffcoreset},
assume that $(f(x) - \sum_{j=0}^r c_j x^j)^2$ is a $L$-Lipschitz function.\footnote{{
For some constant $L>0$, a function $f$ is called $L$-Lipschitz if 
$|f(x)-f(y)|\leq L|x-y|$ for all $x,y$.}}
Then, it holds
\begin{align} \label{eq:coresetbound}
&\mathbf{E}\left[\norm{f^{\odot}(UV^\top) - \Gamma}_2^2 \right]   \nonumber\\
&\leq \frac{2 }{\left(1 + \frac{m \overline{\sigma}^2}{r{C}}\right)}  \norm{\overline{\f}}_2^2 
 + 2 \varepsilon L  \left(\sum_{i=1}^n \norm{\v_i}_2\right),
\end{align}
where $\Gamma$ is the output of Algorithm \ref{alg:polyts} with $\c$, $\v_i$
is the $i$-th row of $V$,
$\overline{\sigma}$ is the smallest singular value of $D^{1/2}\overline{X}$ and 
${C}:=\max ( 5 \|{U}\|_F^2 \|V\|_F^2, (2+3^r)  (\sum_j (\sum_{k} {U}_{jk}^{2})^r)(\sum_j (\sum_{k} V_{jk}^{2})^r))$.
\end{theorem}

{
The proof of Theorem \ref{thm:coreset} is given in the supplementary material.
Observe that when $\varepsilon$ is small, the error bound \eqref{eq:coresetbound}
becomes closer to that under the optimal coefficient \eqref{eq:errbound3}.
To find a coreset with small $\varepsilon$,
we use the greedy $k$-center algorithm \cite{har2008geometric} described in
Algorithm \ref{alg:kcenter}.
We remark that the greedy $k$-center runs in $\bigo{ndk}$ time
and it marginally increases the overall complexity of \PTS in our experiments.
Moreover, we indeed observe that 
various real-world datasets used in our experiments
are well-clustered, which leads to a small $\varepsilon$ 
(see supplementary material for details).
}

{
Algorithm \ref{alg:coreset} summarizes the proposed scheme
for computing the approximated coefficient vector $\c$ in \eqref{eq:coeffcoreset} with
the greedy $k$-center algorithm.
Here,
we run the greedy $k$-center on both rows in $U$ and $V$,
and choose the one with a smaller value in the second term in \eqref{eq:coresetbound}, i.e., 
$\varepsilon_U \sum_{i} \norm{\v_i}_2$ or $\varepsilon_V \sum_{i} \norm{\u_i}_2$
(see also line 4-8 in Algorithm \ref{alg:coreset}).
We remark that 
%
Algorithm \ref{alg:coreset} requires $\bigo{nk(d+r^2) + r^3}$ operations, 
hence, applying this to \PTS results in $\bigo{nk(d+r^2) + r^3} + \bigo{nr(d + m \log m)}$ time in total.
If one chooses $m, k,r=\bigo{1}$, the overall running time becomes 
$\bigo{nd}$ which is linear in the size of input matrix.
For example, we choose $r, k, m=10$ in our experiments.

}

{\bf Chebyshev polynomial regression for avoiding a numerical issue.}
Recall that $X$ contains $(UV^\top)_{ij}^r$ (see Definition \ref{def:not}).
If entries in $UV^\top$ are greater (or smaller) than $1$ and degree $r$ is 
large, $X$ can have huge (or negligible) values.
This can cause a numerical issue for computing the optimal coefficient
\eqref{eq:optc} (or \eqref{eq:coeffcoreset}) using $X$.
To alleviate the issue, 
we suggest to construct a matrix $X^\prime \in \R^{n^2 \times (r+1)}$ whose 
entries are the output of the Chebyshev polynomials:
$(X^\prime)_{ij} = t_j( (UV^\top)_{k\ell})$ 
where $k, \ell \in [n]$ corresponds to index $i \in [n^2]$, $j \in 
[r+1]$
and 
$t_j(x)$ is the Chebyshev polynomial of degree $j$ \cite{mason2002chebyshev}.
Now, the value of $t_j(x)$ is always in $[-1,1]$ 
and does not monotonically increase or decrease with respect to the degree $j$.
Then, we find the optimal coefficient $\c^\prime \in \R^{r+1}$ based 
on Chebyshev polynomials as follows:
\begin{align}
{\c^{\prime}} = 
\left( 
{X^{\prime}}^\top  X^{\prime} + R^\top W^\top W R
\right)^{-1} 
{X^{\prime}}^\top \f,
\end{align}
where 
$R \in \R^{(r+1) \times (r+1)}$ satisfies that
$\c^* = R\c^\prime$ and can be easily computed.
Specifically, it converts $\c^\prime \in \R^{r+1}$ into the coefficient
based on monomials, i.e.,
$\sum_{j=0}^r c_j^* x^j = \sum_{j=0}^r c_j^\prime t_j(x)$.  
We finally remark that to find $t_j$,
one needs to find a closed interval containing all $(UV^\top)_{ij}$.
To this end, we use the interval $[-a,a]$ where 
$a = \left(\max_{i} \|\u_i\|_2\right) \left(  \max_{j} \|\v_j\|_2\right)$ and $\u_i$ is the $i$-row of the matrix $U$. 
It takes $\bigo{nd}$ time and contributes marginally to the overall complexity of {\sc Poly-TensorSketch}.

%% file: exp.tex
In this section, we report the empirical results of \PTS for the element-wise 
matrix functions
under various machine learning applications.\footnote{The datasets used in Section \ref{sec:exp1} and \ref{sec:exp2} are available at {\url{http://www.csie.ntu.edu.tw/~cjlin/libsvmtools/datasets/} 
{and \url{http://archive.ics.uci.edu/}}}.}
All results are reported by averaging over $100$ and $10$ independent trials 
for the experiments in Section \ref{sec:exp1} and those in other sections, 
respectively.
Our implementation and experiments are available at 
\url{https://github.com/insuhan/polytensorsketch}.

\setlength{\dbltextfloatsep}{12pt}
\begin{figure*}[ht]
\vspace{-0.1in}
\centering
\subfigure[synthetic dataset]{
\includegraphics[width=0.24\textwidth]{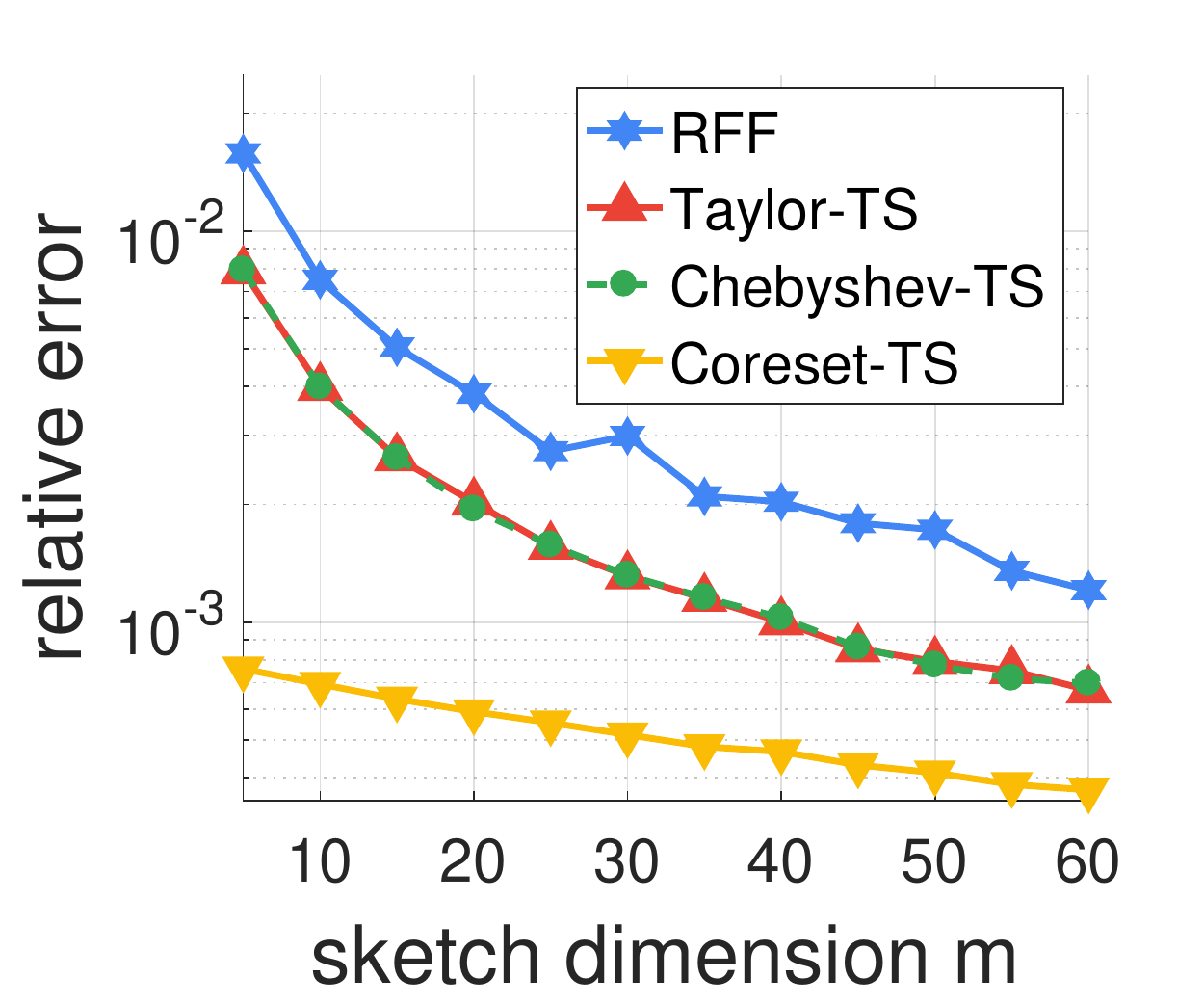}
\includegraphics[width=0.24\textwidth]{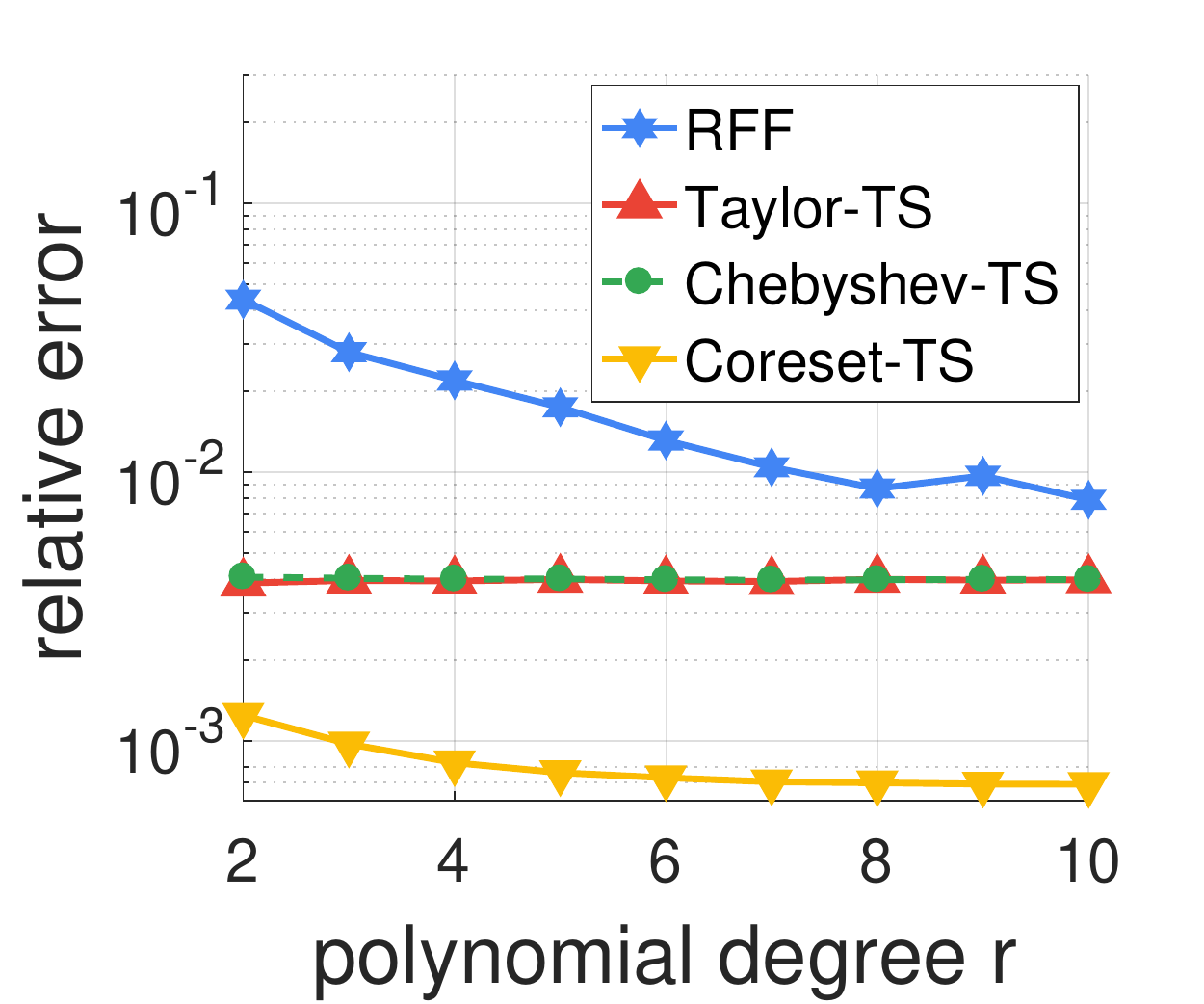}
\includegraphics[width=0.24\textwidth]{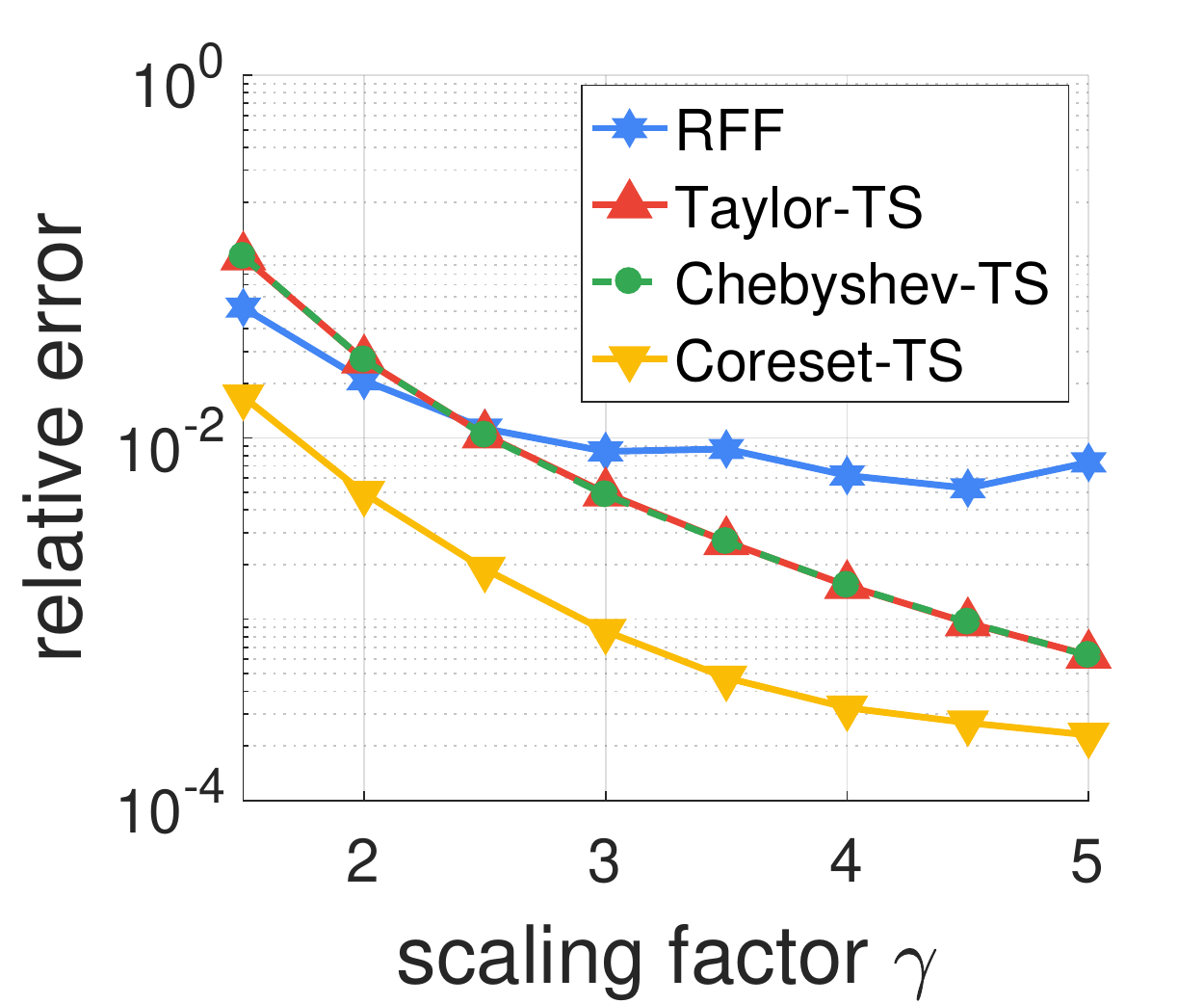}
\includegraphics[width=0.24\textwidth]{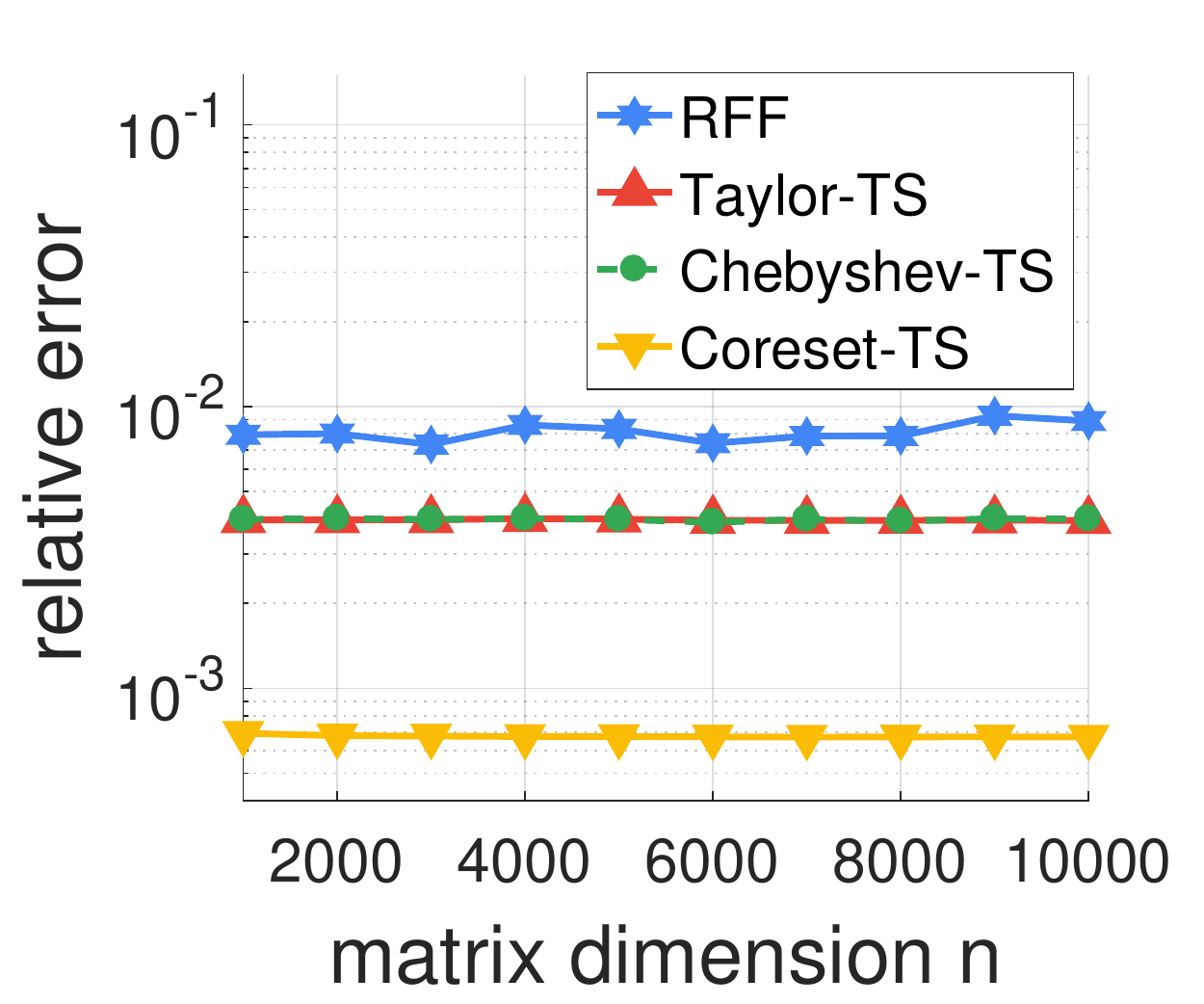}
}
\\
\vspace{-0.15in}
\subfigure[\segment dataset]{
\centering
\includegraphics[width=0.24\textwidth]{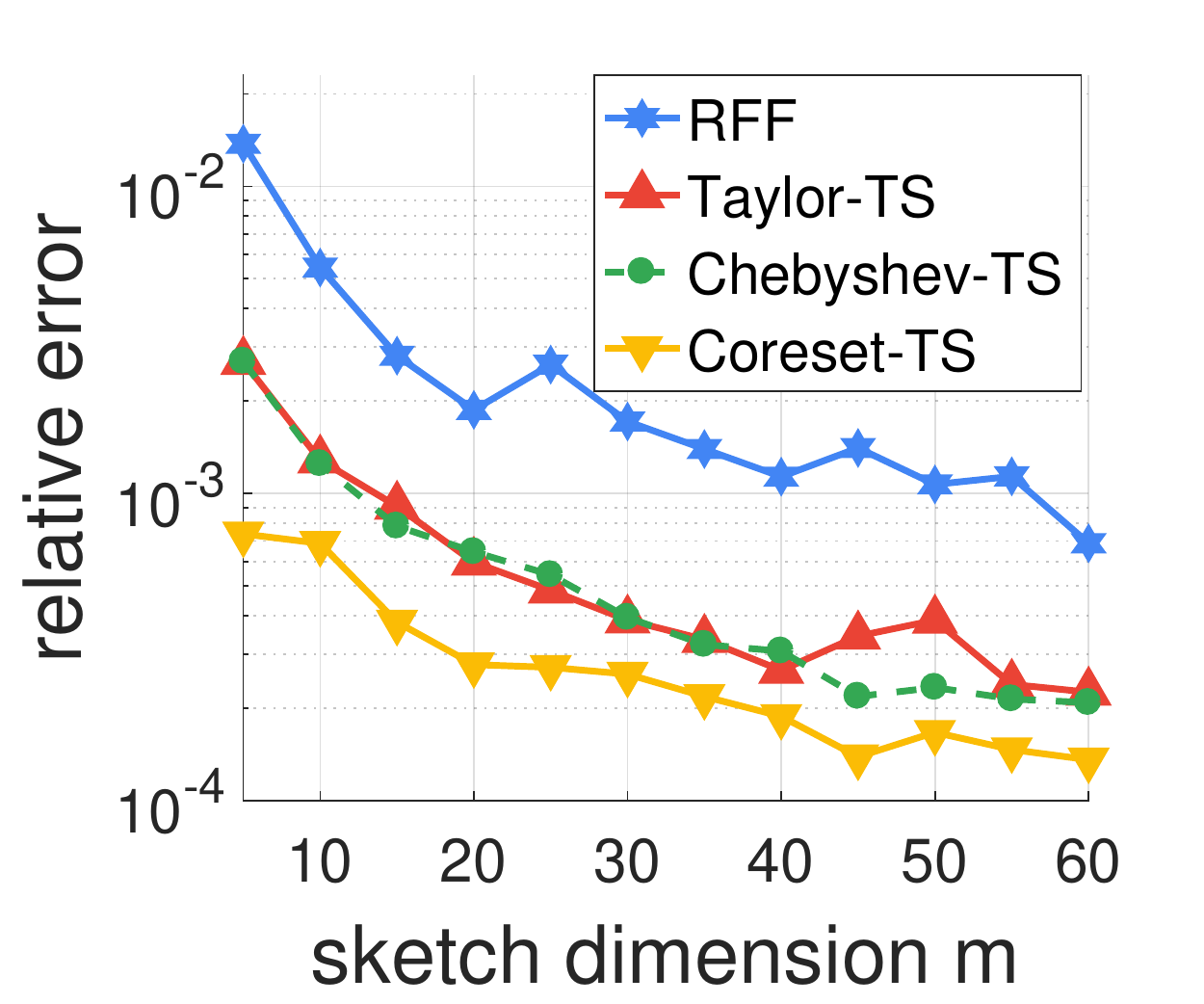}
\includegraphics[width=0.24\textwidth]{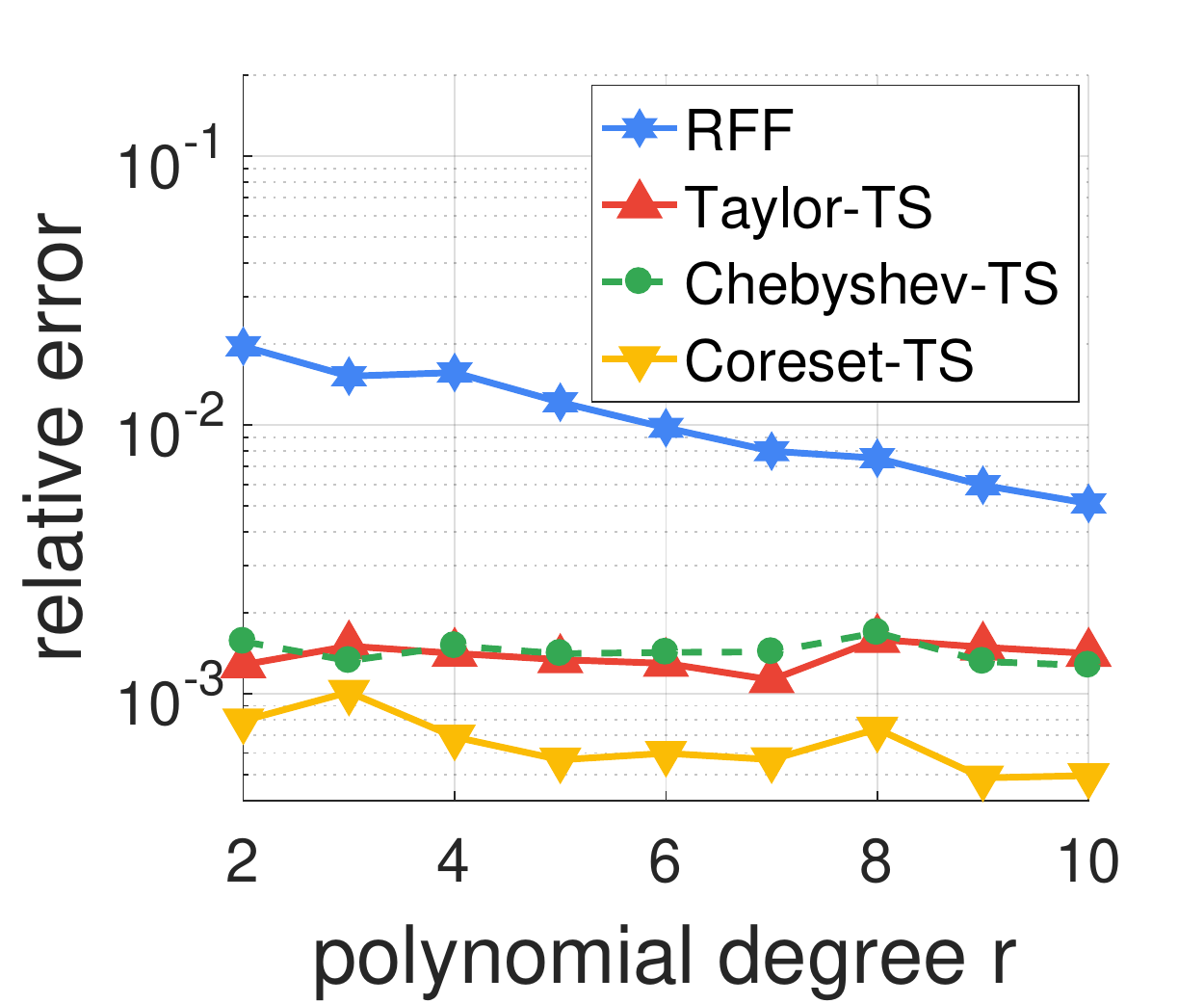}
}
\subfigure[\usps dataset]{
\centering
\includegraphics[width=0.24\textwidth]{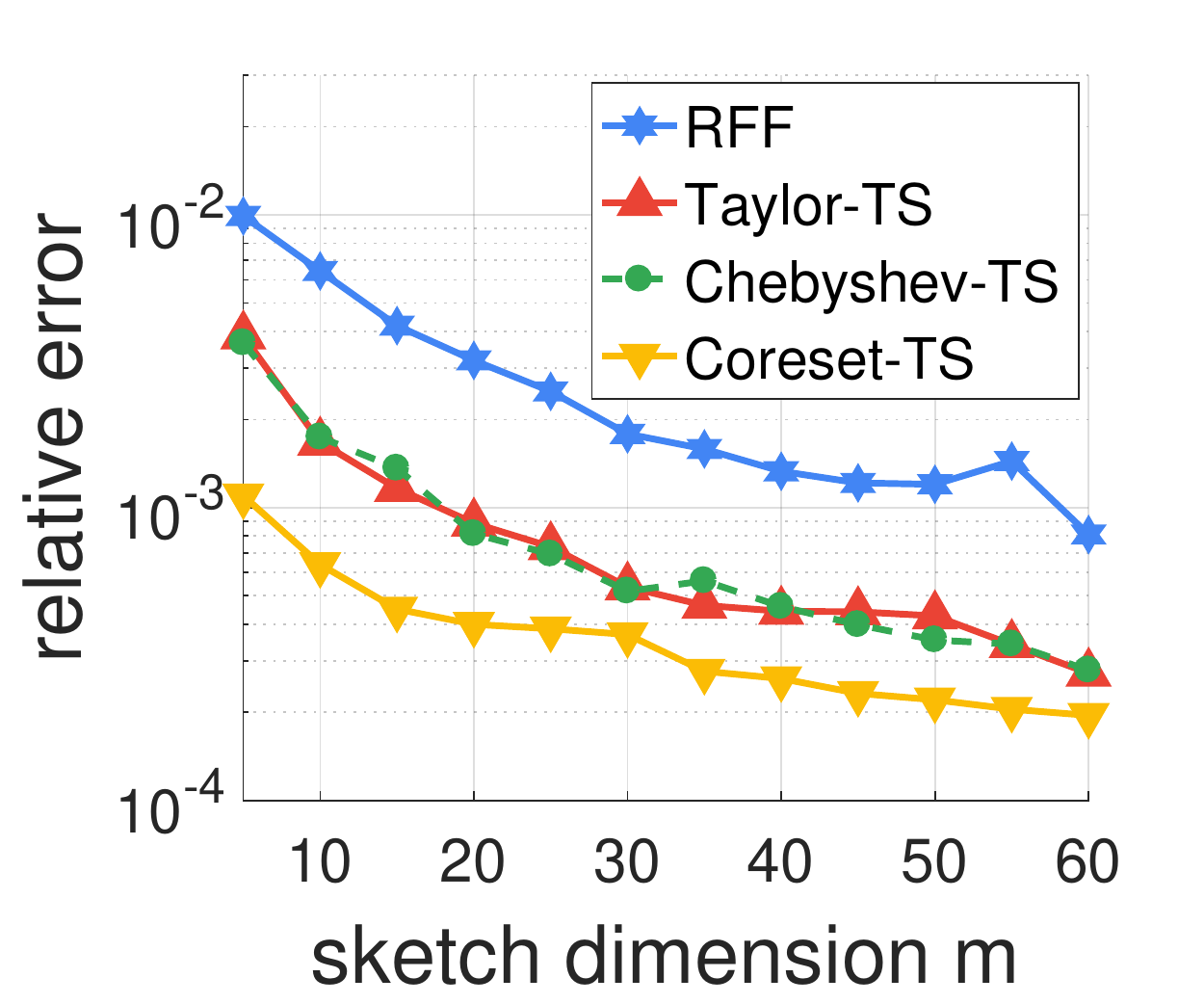}
\includegraphics[width=0.24\textwidth]{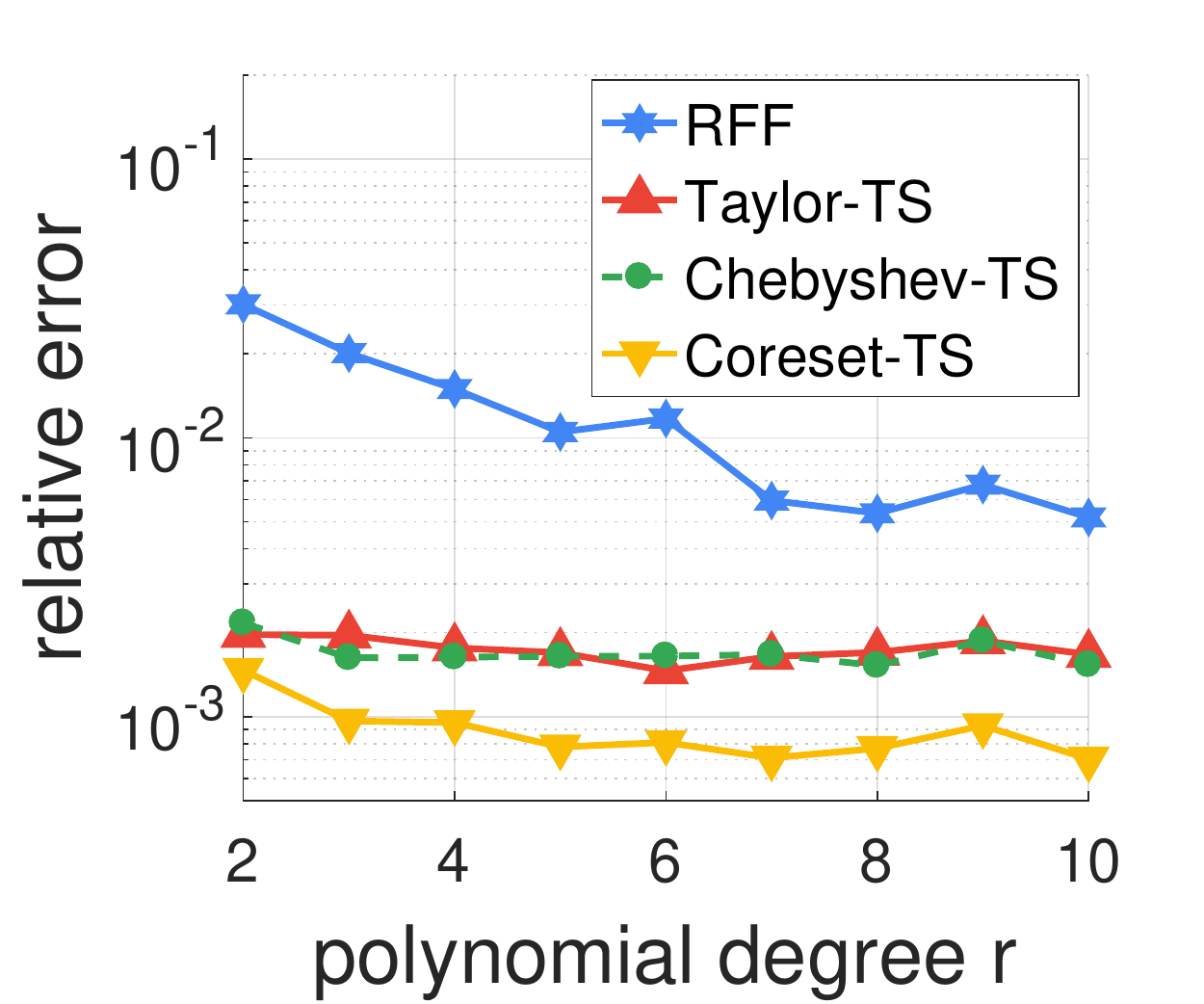}
}
\vspace{-0.1in}
\caption{Kernel approximation under (a) synthetic, (b) \segment and (c) \usps 
dataset.
We set sketch dimension $m=10$, polynomial degree $r = 10$
and coreset size $k=10$ unless stated otherwise. All test methods have 
comparable running times.
}\label{fig:kernelapprox}
\end{figure*}
\begin{table*}[ht] 
\caption{Classification error, kernel approximation error and speedup for 
classification (training time) with kernel {\sc SVM} under various real-world 
datasets.} 
\label{table:cls}
\vspace{-0.1in}
\begin{center}
\setlength{\tabcolsep}{10pt}
\def\arraystretch{1.1}
\begin{small}
\begin{tabular}{ccccccccc}
\toprule
\multirow{2}{*}{Dataset} & \multicolumn{2}{c}{Statistics} & 
\multicolumn{3}{c}{Classification error (\%)} & \multicolumn{2}{c}{Kernel 
approximation error} & {Speedup} \\ \cmidrule{2-9}
                 &  $n$    & $d$   & Exact  & {\sc RFF}     & Coreset-TS     
                 &   {\sc RFF}   & Coreset-TS  & Coreset-TS       \\ \midrule
\segment & $2{,}310$ & $19$   
& $3.20$ & $3.28$ & $3.22$ & $1.7\times10^{-3}$ & $3.21\times 10^{-4}$ & $8.26$ 
\\
\satimage& $4{,}335$ & $36$ 
& $23.54$ & $24.11$ & $23.74$  &  $6.5\times 10^{-3}$ & $3.54\times 10^{-3}$ & 
$10.06$ \\
\anuran& $7{,}195$ & $22$ 
& $37.26$ & $40.33$ & $37.36$ & $6.1\times 10^{-3}$ & $3.1\times 10^{-4}$ & 
$49.45$  \\
\usps& $7{,}291$ & $256$  & $2.11$ & $6.50$  & $5.36$ & $1.2\times10^{-3}$& 
$1.61\times 10^{-4}$ & $27.15$ \\
\grid& $10{,}000$ & $12$ & $3.66$ & $18.27$ & $15.99$ & $2.34\times10^{-3}$& 
$1.02\times10^{-3}$ & $4.23$ \\
\mapping& $10{,}845$ & $28$ & $15.04$ & $18.57$ & $17.35$ & 
$6.09\times10^{-3}$& $2.84\times10^{-4}$ & $4.17$ \\
\letter  & $20{,}000$ & $16$  & $2.62$ & $11.38$  & $10.30$ & $3.06 \times 
10^{3}$  & $1.29$ & $40.60$ \\ \bottomrule
\end{tabular}
\end{small}
\end{center}
\end{table*}
\vspace{-0.1in}

\subsection{{\sc RBF} Kernel Approximation} \label{sec:exp1}
Given $U = [\u_1, \dots, \u_n]^\top \in \R^{n \times d}$, 
the {\sc RBF} kernel $K=[K_{ij}]$ is defined as 
$K_{ij} := \exp( -\| \u_i - \u_j \|_2^2 / \gamma)$ for $i,j \in [n]$ and 
$\gamma > 0$.
It can be represented using the element-wise matrix exponential function:
\begin{align}
K = Z \exp^{\odot}(2 UU^\top / \gamma) Z,
\end{align}
where $Z$ is the $n$-by-$n$ diagonal matrix with 
$Z_{ii}=\exp(-\norm{\u_i}_2^2/\gamma)$.
One can approximate the element-wise matrix function $\exp^{\odot}(2 UU^\top / 
\gamma) \approx \Gamma$ 
where $\Gamma$ is the output of {\sc Poly-TensorSketch} with $f(x) = \exp(2 x / 
\gamma)$ and $K \approx Z \Gamma Z$.
The {\sc RBF} kernel has been used in many applications including 
classification~\cite{pennington2015spherical}, covariance 
estimation~\cite{wang2015beyond}, 
Gaussian process~\cite{rasmussen2003gaussian}, determinantal point 
processes~\cite{affandi2014learning} where they commonly require 
multiplications between the kernel matrix and vectors.

For synthetic kernels, we generate random matrices $U \in \mathbb{R}^{1{,}000 
\times 50}$ whose entries are drawn from the normal distribution 
$\mathcal{N}(0,1/{50})$.
For real-world kernels, we use \segment and \usps datasets.
We report the average of relative error for $n^2$ entries of $K$ under varying 
parameters, 
i.e., sketch dimension $m$ and polynomial degree $r$. 
We choose $m=10$, $r=10$ 
and $k=10$ as the default configuration.
We compare our algorithm, denoted by {Coreset-TS, (i.e., {\sc 
Poly-TensorSketch} 
using the coefficient computed as described in Section \ref{sec:coreset})}
with the random Fourier feature ({\sc RFF}) \cite{rahimi2008random} of the same 
computational complexity, i.e., its embedding dimension is chosen to $mr$
so that the rank of their approximated kernels is the same and their running 
times are comparable.
We also benchmark {\sc Poly-TensorSketch} using coefficients from Taylor and 
Chebyshev series expansions which are denoted by Taylor-TS and Chebyshev-TS, 
respectively. 
As reported in Figure \ref{fig:kernelapprox}, 
we observe that Coreset-TS consistently outperforms the competitors under all 
tested settings and datasets,
where its error is often at orders of magnitude smaller than that of 
{\sc RFF}.
In particular, observe that
the error of Coreset-TS tends to decreases with respect to the polynomial 
degree $r$,
which is not the case for Taylor-TS and Chebyshev-TS (suboptimal versions of 
our algorithm).
We additionally perform the \PTS using the optimal coefficient and compare it 
with Coreset-TS in the supplementary material.

\begin{figure*}[t]
\begin{center}
\centering
\subfigure[]{
\includegraphics[width=0.26\textwidth]{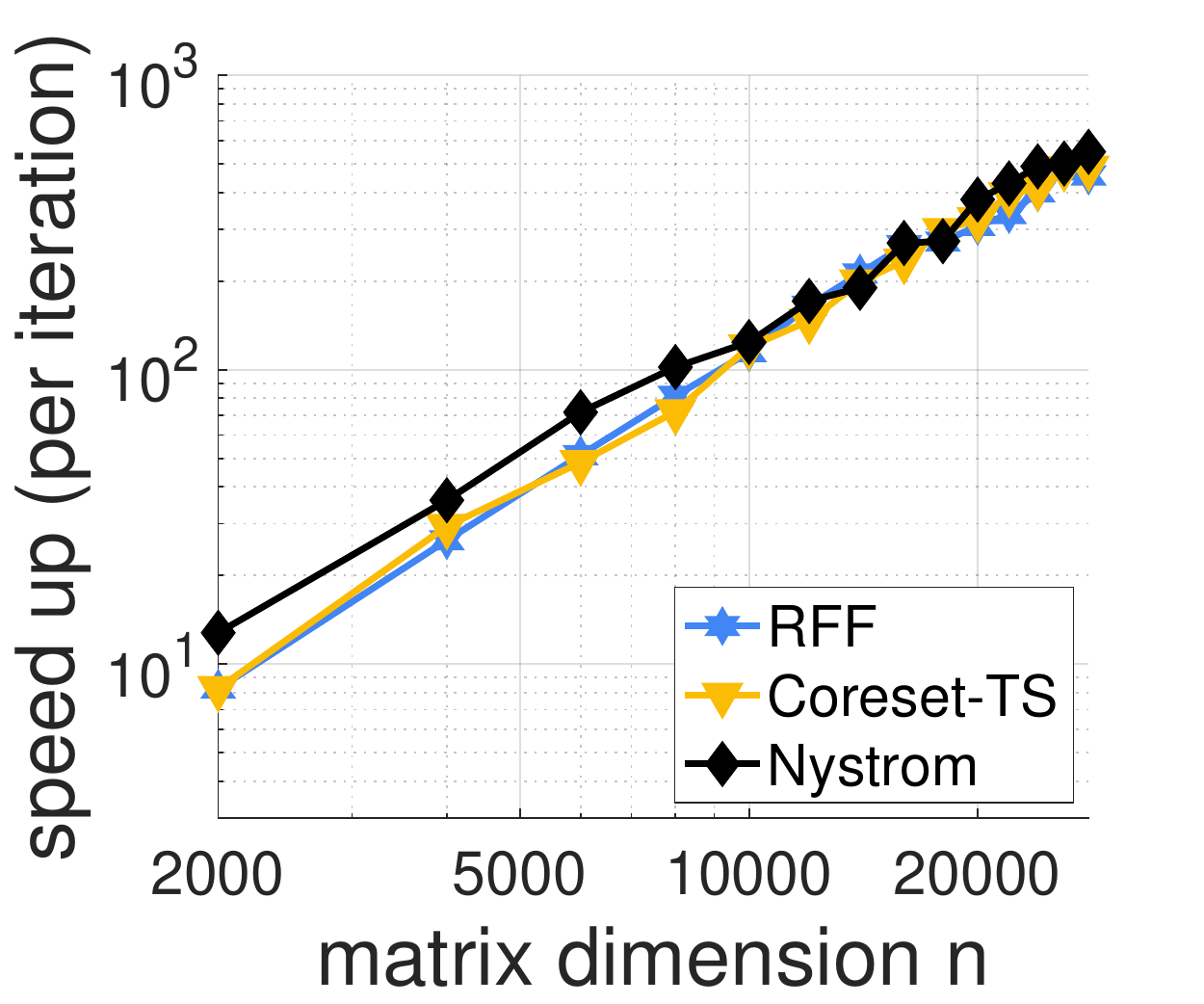}
}
\hspace{-0.15in}
\subfigure[]{
\includegraphics[width=0.26\textwidth]{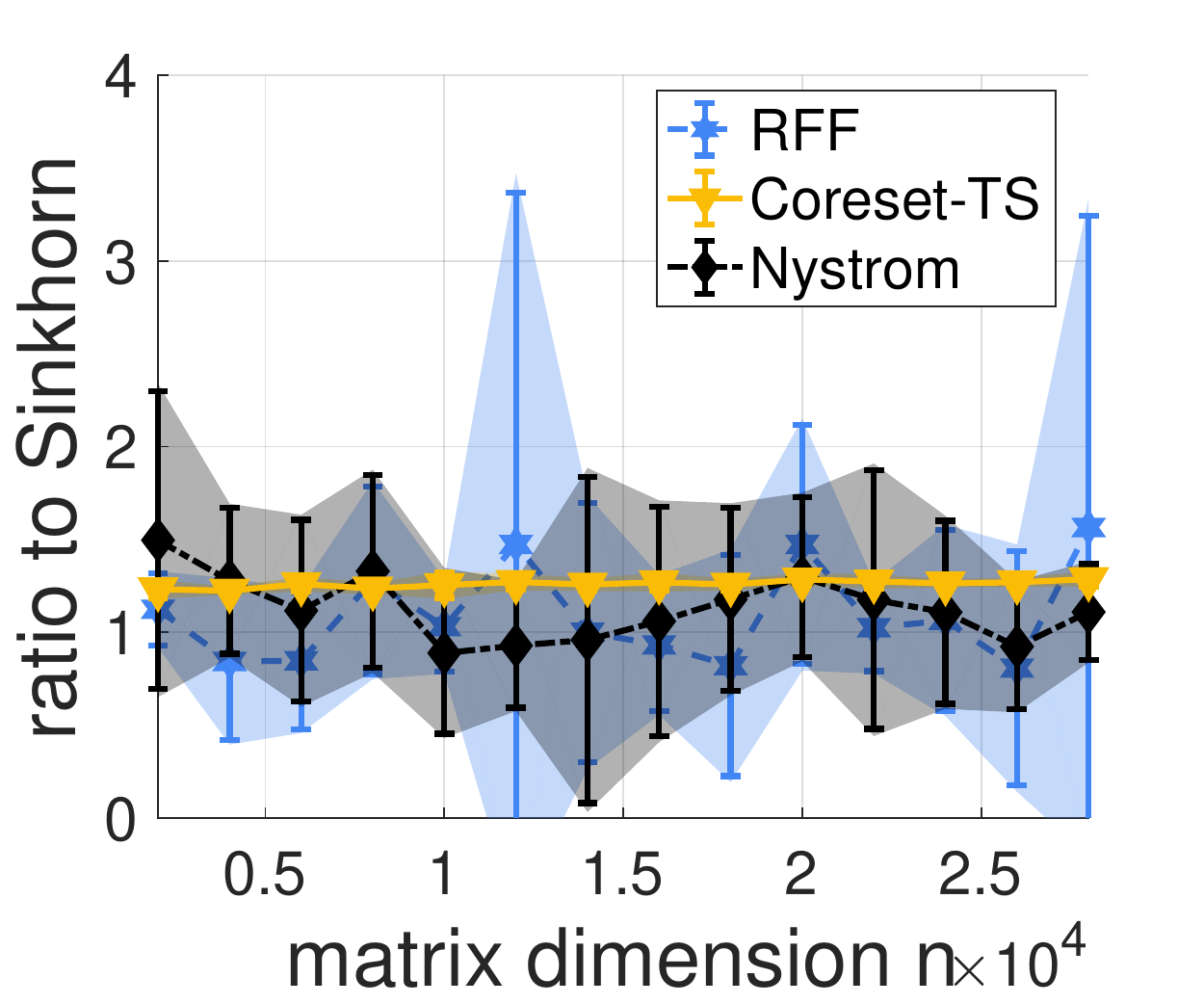}
}
\hspace{0.5in}
\subfigure[]{
\hspace{-0.1in}
\centering
%
%
%
\includegraphics[height=0.17\textheight]{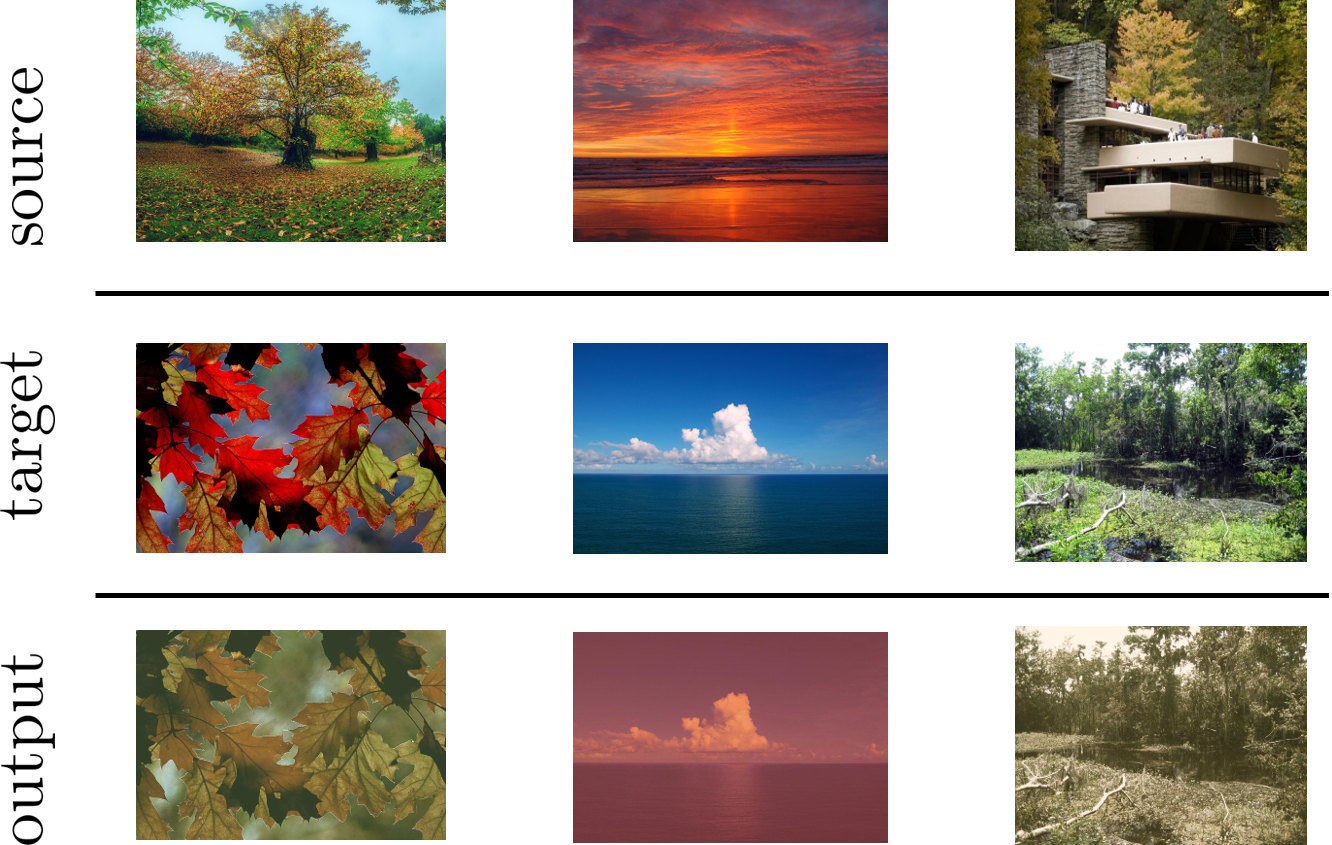}
}
\vspace{-0.15in}
\end{center}
\caption{
Speedup per iteration (a) and the ratio to Sinkhorn objective (b) of our 
algorithm, {\sc RFF} and Nystrom approximation 
averaged over $3$ tested images in (c). The Sinkhorn algorithm transforms the 
color information of source images to targets.
}\label{fig:sink}
\end{figure*}

\subsection{Classification with Kernel {\sc SVM}} \label{sec:exp2}

Next, we aim for applying our algorithm (Coreset-TS) to classification tasks 
using the support vector machine ({\sc SVM}) based on {\sc RBF} kernel.
Given the input data $U \in \R^{n \times d}$, 
our algorithm can find a feature $T_U \in \R^{n \times m^\prime}$ such that $K 
\approx T_U T_U^\top$ where $K$ is the {\sc RBF} kernel of $U$ 
(see Section \ref{sec:exp1}). 
One can expect that a linear {\sc SVM} using $T_U$ 
shows a similar performance compared to the kernel {\sc SVM} using $K$.
However, 
for $m^\prime \ll n$,
the complexity of a linear {\sc SVM} is much cheaper than that of the kernel 
method both for training and testing.
In order to utilize our algorithm, 
we construct $T_U = \left[\sqrt{c_0} T_U^{(0)} \dots \sqrt{c_r} T_U^{(r)} 
\right]$ 
where $T_U^{(0)}, \dots, T_U^{(r)}$ are the {\sc TensorSketch}s of $U$ in 
Algorithm \ref{alg:polyts}. 
Here, the coefficient $c_j$ should be positive and one can compute the optimal 
coefficient \eqref{eq:optc} by adding non-negativity condition, which is 
solvable using simple quadratic programming with marginal additional cost. 


We use the open-source {\sc SVM} package ({\sc LIBSVM}) \cite{CC01a} and 
compare our algorithm with 
the kernel {\sc SVM} using the exact {\sc RBF}
and the linear {\sc SVM} using the embedded feature from {\sc RFF} of the same 
running time with ours.
We run all experiments with $10$ cross-validations and report the average of 
the classification error on the validation dataset.
{We set $m=20$ for the dimension of sketches and
$r=3$ for the degree of the polynomial.
}
Table \ref{table:cls} summarizes the results
of classification errors, kernel approximation errors and speedups of our 
algorithm under various real-world datasets.
Ours (Coreset-TS) shows better classification errors compared to {\sc RFF} of 
comparable running time and runs up to $50$ times faster than the exact method. 
We also observe that large $m$ can improve the performance 
(see the supplementary material).

\subsection{Sinkhorn Algorithm for Optimal Transport} \label{sec:exp3}
We apply Coreset-TS to 
the Sinkhorn algorithm for computing the optimal transport distance 
\cite{cuturi2013sinkhorn}.
The algorithm is the entropic regularization for approximating the distance of 
two discrete probability density.
It is a fixed-point algorithm where each iteration requires multiplication of 
an element-wise matrix exponential function with a vector. 
More formally, given $\a, \b\in \R^{n}$, 
and two distinct data points of the same dimension
$\{\x_i\}_{i=1}^{n}$, $\{\y_j\}_{j=1}^{n}$,
the algorithm iteratively computes 
$\u = \a \oslash \left(\exp^{\odot}( - \gamma D) \v\right)$ and 
$\v = \b \oslash \left(\exp^{\odot}( - \gamma D)^\top \u\right)$ for some 
initial $\v$ 
where 
$D_{ij} := \|\x_i - \y_j\|_2^2$ and 
$\oslash$ denotes the element-wise division.
Hence, the computation can be efficiently approximated using our algorithm or 
{\sc RFF}.
{
We also perform 
the Nystr{\"o}m approximation \cite{williams2001using}
which was
recently used for developing a scalable Sinkhorn algorithm 
\cite{altschuler2019massively}. 
We provide all the details in the supplementary materials.
}

\setcounter{footnote}{0}

Given a pair of source and target images as shown in Figure \ref{fig:sink} 
(c),\footnote{Images from \url{https://github.com/rflamary/POT}.} %
we randomly sample $\{\x_i\}_{i=1}^{n}$ from RGB pixels in the source image 
and $\{\y_j\}_{j=1}^{n}$ from those in the target image. 
We set $m=20, d = 3, r = 3$ and $\gamma = 1$. 
Figure \ref{fig:sink} (a) reports the speedup per iteration of the tested 
approximation algorithms over the exact computation
and Figure \ref{fig:sink} (b) shows the ratios the objective value of the 
approximated Sinkhorn algorithm to the exact one after $10$ iterations.
As reported in Figure \ref{fig:sink} (a), all algorithms run at orders of 
magnitude faster than the exact Sinkhorn algorithm.
Furthermore, the approximation ratio of ours is much more stable, while both 
{\sc RFF} and Nystr{\"o}m have huge variances without any tendency on the 
dimension $n$. 

%
%

%
%

\ifdefined\isaccepted
    \begin{table}[t] 
    \caption{Test errors on CIFAR100 of linearization of $2$ fully-connected layers of AlexNet, 
    i.e., $W_2 \mathtt{sigmoid}^{\odot}(W_1 \x) \approx (W_2 T_{W_1}) T_{\x}$ where 
    both the input and hidden dimensions are $d=d_h=1{,}024$ and the degree $r=3$.
    } \label{table:nn}
    \vspace{-0.10in}
    \begin{center}
    \begin{small}
    \setlength{\tabcolsep}{3pt}
    \def\arraystretch{1.0}
    \begin{tabular}{cC{3cm}cc}
    \toprule
    Model & \shortstack{Complexity} & \shortstack{$m$} & \shortstack{Error ($\%$)}  
    \\ \midrule
    $W \x$ & $\bigo{d}$ & $-$ & $45.67$  \\ \cmidrule{1-4}
    \multirow{4}{*}{$(W_2 T_{W_1}) T_{\x}$} & 
    \multirow{4}{*}{${\mathcal{O}}(r(d + m \log m))$} & $100$ & 
    $53.09$ \\ 
    & &$200$ & $47.18$ \\
    & &$400$ & $43.90$ \\
    & &$800$ & $41.73$ \\ 
    \cmidrule{1-4}
    $W_2 \mathtt{sigmoid}^{\odot}(W_1 \x)$ & $\bigo{d_h d}$ & $-$ & $39.02$ \\
    \bottomrule
    \end{tabular}
    \end{small}
    \end{center}
    \vspace{-0.25in}
    \end{table}
\else
    \begin{table}[t] 
    \caption{Test errors on MNIST of linearization of $2$-layer neural network, 
    i.e., $W_2 \mathtt{sigmoid}^{\odot}(W_1 \x) \approx (W_2 T_{W_1}) T_{\x}$ where
    the input and hidden dimensions are $d=784$, $d_h=800$, respectively, and the 
    degree $r=3$.
    } \label{table:nn}
    \vspace{-0.15in}
    \begin{center}
    \begin{small}
    \setlength{\tabcolsep}{3pt}
    \def\arraystretch{1.0}
    \begin{tabular}{cC{3cm}cc}
    \toprule
    Model & \shortstack{Complexity} & \shortstack{$m$} & \shortstack{Error ($\%$)}  
    \\ \midrule
    $W \x$ & $\bigo{d}$ & $-$ & $9.13$  \\ \cmidrule{1-4}
    \multirow{4}{*}{$(W_2 T_{W_1}) T_{\x}$} & 
    \multirow{4}{*}{${\mathcal{O}}(r(d + m \log m))$} & $100$ & 
    $11.45$ \\ 
    & &$200$ & $9.07$ \\
    & &$400$ & $8.08$ \\
    & &$800$ & $7.62$ \\ 
    \cmidrule{1-4}
    $W_2 \mathtt{sigmoid}^{\odot}(W_1 \x)$ & $\bigo{d_h d}$ & $-$ & $2.29$ \\
    \bottomrule
    \end{tabular}
    \end{small}
    \end{center}
    \vspace{-0.25in}
    \end{table}
\fi

\subsection{Linearization of Neural Networks} \label{sec:exp4}
Finally, we demonstrate that our method has a potential to obtain a low-complexity 
model by linearization of a neural network. To this end, we consider $2$ 
fully-connected layers in AlexNet~\cite{krizhevsky2012imagenet}
where each has $1{,}024$ hidden nodes and it is trained on CIFAR100 
dataset~\cite{krizhevsky2009learning}.
Formally, given an input $\x \in \R^{1{,}024}$, its predictive class corresponds to 
$\arg \max_i \left[W_2 \  \mathtt{sigmoid}^{\odot}(W_1 \x)\right]_i$ 
where $W_1 \in \R^{1{,}024 \times 1{,}024}, W_2 \in \R^{1{,}024 \times 100}$ 
are model parameters. We first train the model for $300$ epochs using ADAM optimizer 
\cite{kingma2014adam} with $0.0005$ learning rate. 
Then, we approximate $\mathtt{sigmoid}^{\odot}(W_1 \x) \approx T_{W_1} T_{\x}$ 
for $T_{W_1} \in \R^{800 \times mr}, T_{\x} \in \R^{mr}$ using Coreset-TS.
After that, we fine-tune the parameters $T_{W_1}, W_2$ for $200$ epochs 
and evaluate the final test error as reported in Table \ref{table:nn}.
We choose $r=3$ and explore various $m \in \{100, 200, 400, 800\}$. 
Observe that the obtained model $(W_2 T_{W_1}) T_{\x}$ is linear with respect 
to $T_{\x}$ and its complexity (i.e., inference time) is much smaller than 
that of the original neural network. 
Moreover, by choosing the sketch dimension $m$ appropriately, 
its performance is better than that of the vanilla linear model $W\x$ as 
reported in Table \ref{table:nn}. 
The results shed a broad applicability our generic approximation scheme 
and more exploration on this line would be an interesting direction in the future.

%% file: conclusion.tex
In this paper, we design a fast algorithm for sketching element-wise matrix 
functions.
Our method is based on combining (a) a polynomial approximation with (b) the randomized 
matrix tensor sketch. Our main novelty is on finding the optimal polynomial 
coefficients for minimizing the overall approximation error bound by 
balancing the errors of (a) and (b).
We 
expect that the generic scheme would enjoy a broader usage in the future. 

%% file: extra_exp.tex
\section{Greedy $k$-center Algorithm}
We recall greedy $k$-center algorithm runs in $\bigo{ndk}$ time given $\u_1, \dots, \u_n \in \mathbb{R}^d$
since it computes the distances between the inputs and the selected coreset $k$ times.
The greedy $k$-center was originally studied for set cover problem with $2$-approximation ratio \cite{vazirani2013approximation}.
Beyond the set cover problem, it has been reported to be useful in many clustering applications. Due to its simple yet efficient clustering performance, we adopt this to approximate the optimal coefficient \eqref{eq:optc} (see Section \ref{sec:coreset}).

\section{t-SNE plots of Real-world Datasets}
\begin{figure}[h]
\centering
\subfigure[\segment]{\includegraphics[width=0.30\textwidth]{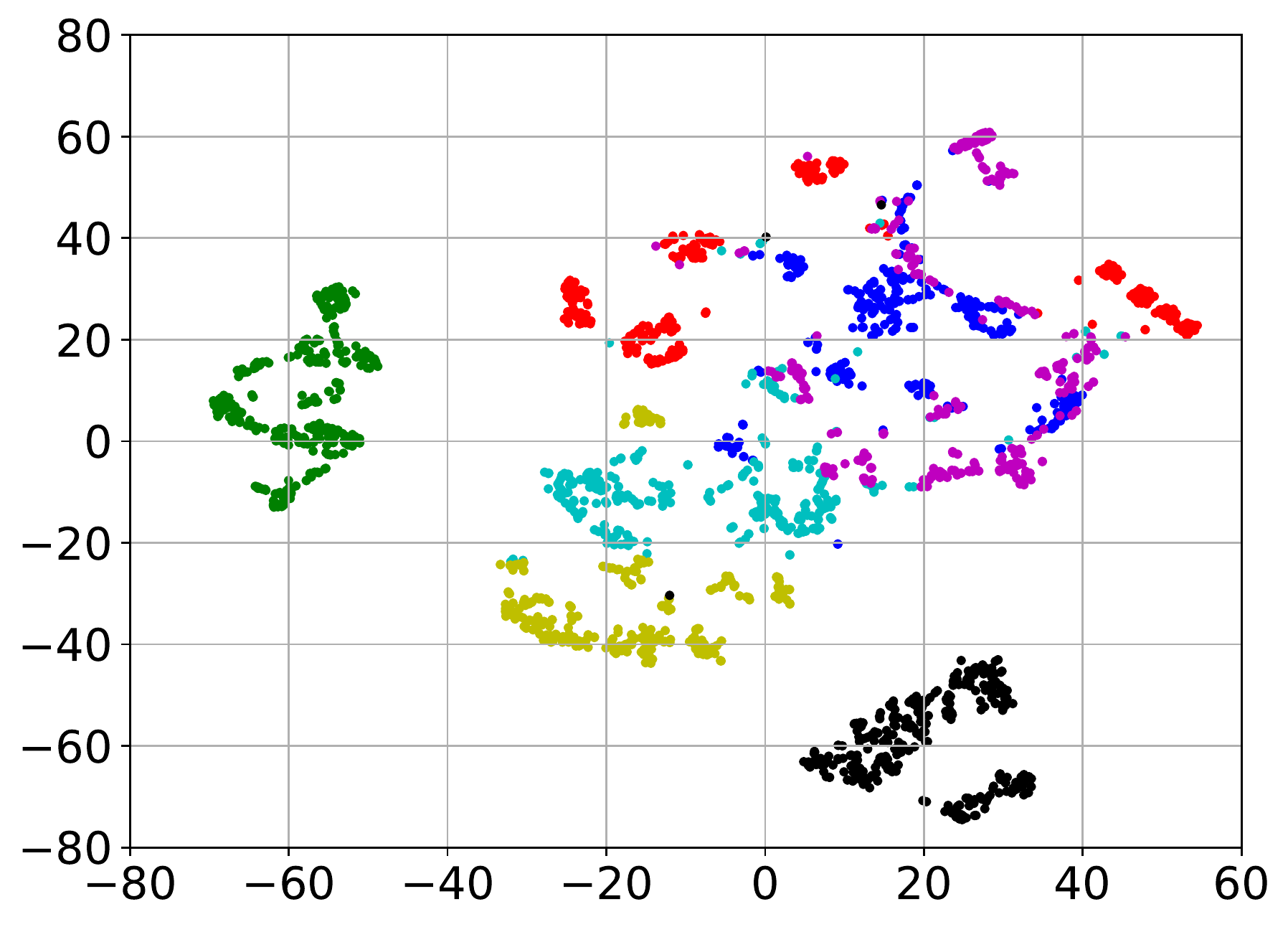}}
\subfigure[\satimage]{\includegraphics[width=0.30\textwidth]{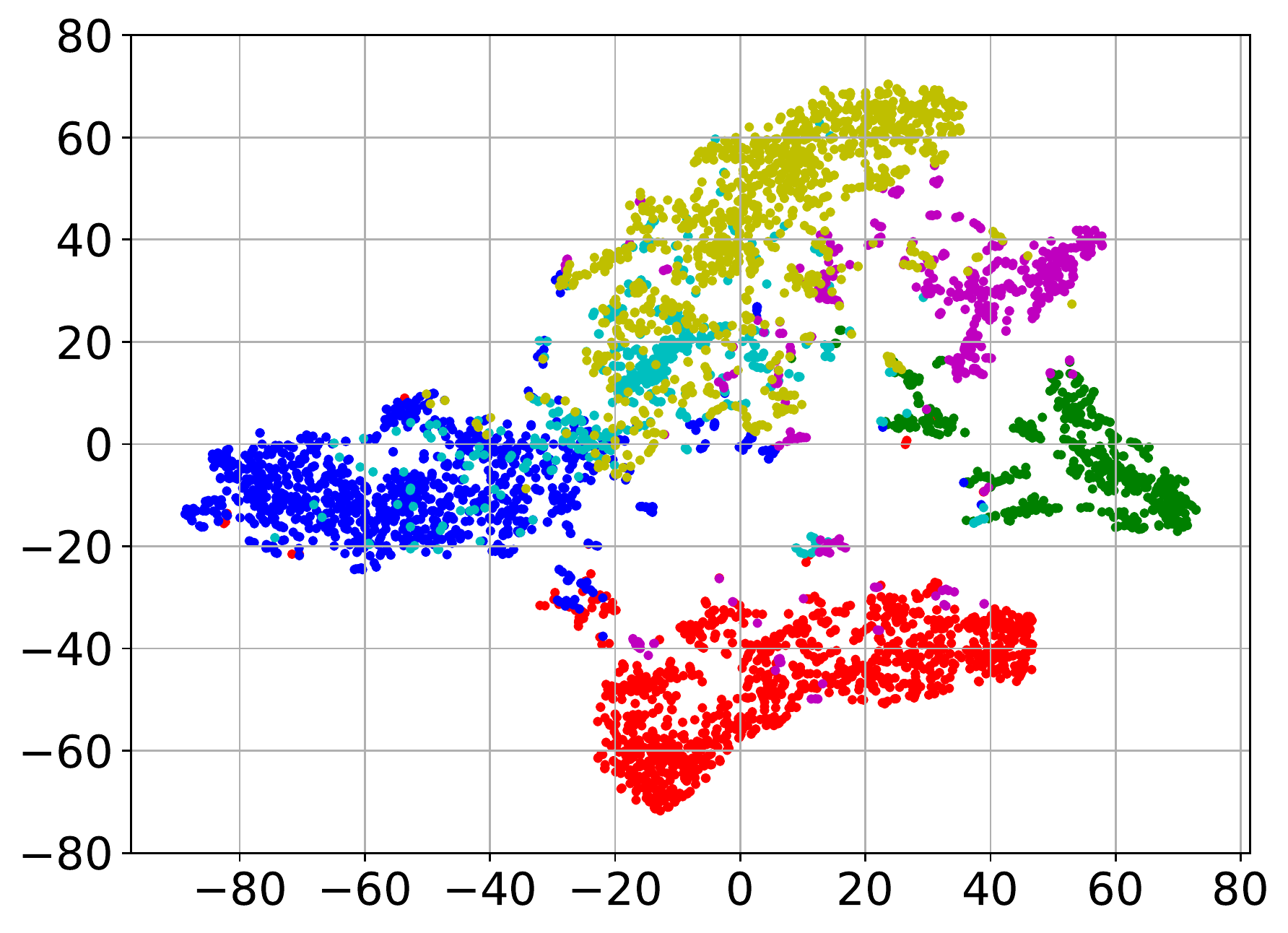}}
\subfigure[\usps]{\includegraphics[width=0.30\textwidth]{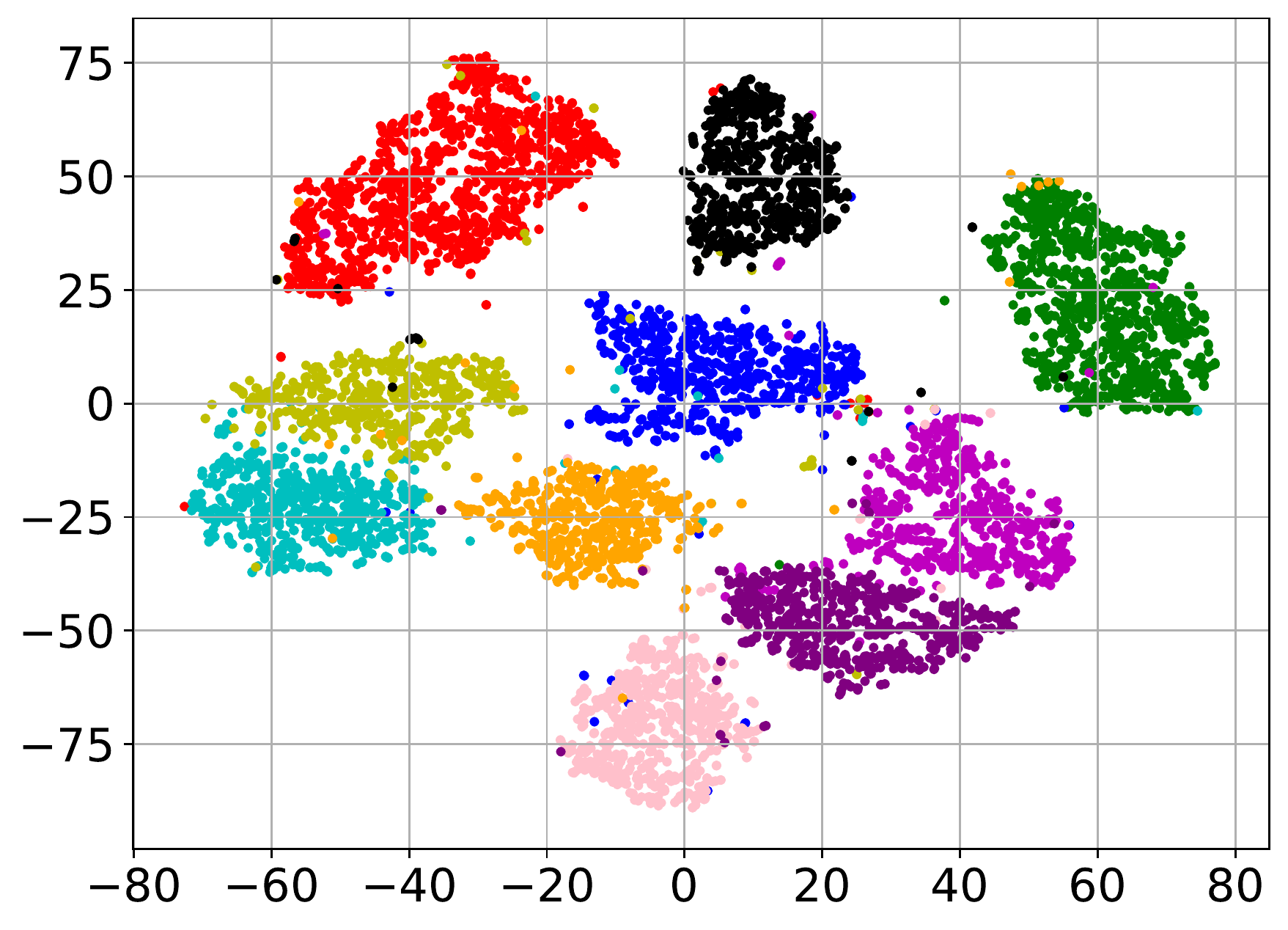}}
\caption{t-SNE plot of (a) \segment, (b) \satimage and (c) \usps datasets. Different colors represent distinguish clusters.}
\label{fig:tsne}
\end{figure}

The error analysis in Theorem \ref{thm:coreset} depends on the upper bound of the distortion of input matrix, i.e., $\varepsilon$ of $\varepsilon$-coreset (see Section \ref{sec:coreset}).
We recall that a smaller $\varepsilon$ gives tighter the error bound (Theorem \ref{thm:coreset}) and this indeed occurs in read-world settings. To verify this, we empirically investigate the t-SNE plot \cite{maaten2008visualizing} for real-world datasets including \segment, \satimage and \usps in Figure \ref{fig:tsne}. Observe that data points in t-SNE are well-clustered and it is expected that the distortion $\varepsilon$ are small.
This can justify to utilize our coreset-based coefficient construction (Algorithm \ref{alg:kcenter} and \ref{alg:coreset}) for real-world problems.

\section{Scalable Sinkhorn Algorithm via Nystr{\"o}m Method}
Recently, \citet{altschuler2019massively} proposed a scalable Sinkhorn algorithm 
based on Nystr{\"o}m approximation. Motivated by their work, we compare our Coreset-TS to the Nystr{\"o}m method as fast Sinkhorn approximations. 
In a nutshell, the Nystr{\"o}m method randomly samples a subset $S \subseteq [n]$ such that $|S|=s$ and approximates the kernel matrix $K \in \R^{n \times n}$ as
\begin{align*}
K ~\approx~ \widetilde{K} := K_{[n],S}(K_{S,S})^{\dagger}K_{S,[n]}
\end{align*}
where $K_{A,B} \in \R^{n \times s}$ is defined as the sub-matrix of $K$ 
whose rows and columns are indexed by $A, B\subseteq [n]$, respectively, and $K^{\dagger}$ is the pseudo-inverse of a matrix $K$.
Observe that given $K_{[n],S}$ and $(K_{S,S})^{\dagger}$
multiplications $\widetilde{K}$ with vectors can be computed efficiently in $\bigo{ns}$ time for $|S|=s$. 
It is easy to check that the approximation takes $\bigo{nds + ns^2 + s^3}$ time 
and remind that ours is $\bigo{nk(d+r^2) + r^3 + nr(d + m \log m)}$. 
Thus, for comparable complexity, we choose $s = \lceil \max( \sqrt{r m}, r) \rceil$ for the Nystr{\"o}m method and report the results in Section \ref{sec:exp3}.

\section{Comparison between Coreset-based Estimation and Optimal Coefficient}
The optimal coefficients $\boldsymbol{c}^*$~\eqref{eq:optc} can provide a smaller approximation error than that computed using the coreset $\boldsymbol{c}^{\prime}$~\eqref{eq:coeffcoreset}, where a larger $k$ (number of clusters) guarantees a smaller gap, i.e., $\norm{\boldsymbol{c}^* - \boldsymbol{c}^\prime}_2$. 
We additionally evaluate the coefficient gap and their kernel approximation errors under the \segment data, 
varying $k$ from $5$ to $30$ as reported in Table~\ref{table:coeffgap}.
Observe that a larger $k$ reduces not only the coefficient gap but also kernel error of the coreset. We use $k=10$ for all reported experiments in our paper since the improvement on the kernel error for $k>10$ tends to be marginal.

\begin{table}[H] 
\caption{
Kernel approximation error between the optimal coefficient and the coreset-based estimation under \segment data. The optimal coefficient achieves the kernel approximation error of $5.17 \times 10^{-4}$ .} \label{table:coeffgap}
\vspace{-0.10in}
\begin{center}
\begin{small}
\setlength{\tabcolsep}{10pt}
\def\arraystretch{1.1}
\begin{tabular}{ccc}
\toprule
$k$ & Coefficient gap & Kernel approximation error by coreset\\ \midrule
$ 5$  &  $3.28\times 10^{-1}$  &  $5.52\times 10^{-4}$\\
$10$  &  $1.72\times 10^{-1}$  &  $5.39\times 10^{-4}$\\
$15$  &  $1.31\times 10^{-1}$  &  $5.33\times 10^{-4}$\\
$20$  &  $1.15\times 10^{-1}$  &  $5.34\times 10^{-4}$\\
$25$  &  $9.07\times 10^{-2}$  &  $5.28\times 10^{-4}$\\
$30$  &  $8.01\times 10^{-2}$  &  $5.23\times 10^{-4}$\\
\bottomrule
\end{tabular}
\end{small}
\end{center}
\end{table}

\section{Classification with Kernel {\sc SVM} using Other Parameters}
In Section \ref{sec:exp2}, we use the sketch dimension $m=20$ but a larger $m$ can reduce the classification error of both Random Fourier Features ({\sc RFF}) and Coreset-TS, while it increases both time and memory complexity. We additionally conduct experiments with $m=50$ under the last $4$ datasets in Table \ref{table:cls} and report the result in Table \ref{table:cls2}.

\begin{table}[H] 
\caption{
Classification error of kernel \textsc{SVM} with $m=20$ and $50$ under $4$ real-world datasets. 
A larger $m$ shows better classification errors for both {\sc RFF} and Coreset-TS.} \label{table:cls2}
\vspace{-0.10in}
\begin{center}
\begin{small}
\setlength{\tabcolsep}{10pt}
\def\arraystretch{1.1}
\begin{tabular}{ccccc}
\toprule
\multirow{3}{*}{Dataset} & \multicolumn{4}{c}{Classification error (\%)} \\ 
\cmidrule{2-5}
                 & \multicolumn{2}{c}{{\sc RFF}} & \multicolumn{2}{c}{Coreset-TS} \\
\cmidrule{2-5}
                 & $m=20$     & $m=50$     & $m=20$     & $m=50$ \\ 
\midrule
\usps            & $6.50$      & $3.55$    & $5.36$     & $3.33$ \\
\grid            & $18.27$     & $11.78$   & $15.99$    & $11.17$ \\
\mapping         & $18.57$     & $17.64$   & $17.35$    & $17.11$ \\
\letter          & $11.38$     & $5.31$    & $10.30$    & $5.13$ \\     
\bottomrule
\end{tabular}
\end{small}
\end{center}
\end{table}

%% file: proof1.tex
\begin{proof}
Let $p_r(x) = \sum_{j=0}^r c_j x^r$ and $[p_r^{\odot}(A)]_{ij} = p_r( A_{ij})$.
Consider the approximation error into the error from (1) polynomial approximation and (2) tensor sketch:
\begin{align*}
\mathbf{E}\left[ \norm{f^{\odot}(UV^\top) - 
\Gamma
}_F^2 \right]
&\leq 
\mathbf{E}\left[
2\norm{f^{\odot}(UV^\top) - p_r^{\odot}(UV^\top)}_F^2
+
2 \norm{p_r^{\odot}(UV^\top) - \Gamma }_F^2
\right] \\
&=
2\norm{f^{\odot}(UV^\top) - p_r^{\odot}(UV^\top)}_F^2
+
2 
\mathbf{E}\left[
\norm{p_r^{\odot}(UV^\top) - \Gamma }_F^2,
\right]
\end{align*}
where the inequality comes from that $(a+b)^2 \leq 2(a^2 + b^2)$ for $a,b \in \mathbb{R}$.
The first error is straightforward from the assumption:
\begin{align*}
\norm{f^{\odot}(UV^\top) - \Gamma}_F^2
&= 
\sum_{i,j} 
\left( 
f\left( (UV^\top)_{ij} \right) - p_r\left( (UV^\top)_{ij} \right)
\right)^2 
\leq n^2 \varepsilon^2
\end{align*}
For the second error, we use Theorem \ref{thm:tensor} to have
\begin{align*}
\mathbf{E}\left[
\left\|
p_r^{\odot}(UV^\top)
- 
\Gamma
\right\|_F^2
\right]
&\leq r \sum_{j=1}^r c_j^2 \ \mathbf{E}\left[
 \left\|(UV^\top)^{\odot j} - T_U^{(j)} {T_V^{(j)}}^\top \right\|_F^2
\right] \\
&\leq 
r 
\sum_{j=1}^r c_j^2 \frac{(2 + 3^j) \left( \sum_{i} (\sum_k U_{ik}^{2})^j \right) \left( \sum_{i} (\sum_k V_{ik}^{2})^j \right)}{m} \\
\end{align*}
Putting all together, we conclude the result and this completes the proof of Proposition \ref{prop:pts}.
\end{proof}

%% file: proof2.tex
\begin{proof}
We recall that 
$
\Gamma = \sum_{j=0}^r c_j T_U^{(j)} T_V^{(j) \top}.
$
and similar to the proof of Proposition \ref{prop:pts} we have
\begin{align*}
\mathbf{E}\left[ \norm{f^{\odot}(UV^\top) - \Gamma}_F^2 \right]
&\leq 
2\norm{f^{\odot}(UV^\top) - p_r^{\odot}(UV^\top)}_F^2
+
2 
\mathbf{E}\left[
\norm{p_r^{\odot}(UV^\top) - \Gamma }_F^2,
\right]
\end{align*}
where the inequality comes from that $(a+b)^2 \leq 2(a^2 + b^2)$.
The error in the first term can be written as
\begin{align*}
\norm{f^{\odot}(UV^\top) - p_r^{\odot}(UV^\top)}_F^2
&= 
\sum_{i,j} 
\left( 
f\left( (UV^\top)_{ij} \right) - p_r\left( (UV^\top)_{ij} \right)
\right)^2 
=
\left \| X \mathbf{c} - \mathbf{f} \right \|_2^2
\end{align*}
where we recall the Definition \ref{def:not}, that is, 
\begin{align*}
X:=
\begin{bmatrix}
    1 & [UV^\top]_{11} & [UV^\top]_{11}^2 & \dots  & [UV^\top]_{11}^r \\
    1 & [UV^\top]_{12} & [UV^\top]_{12}^2 & \dots  & [UV^\top]_{12}^r \\
    \vdots & \vdots & \vdots & \ddots & \vdots \\
    1 & [UV^\top]_{nn} & [UV^\top]_{nn}^2 & \dots  & [UV^\top]_{nn}^r
\end{bmatrix} \in \mathbb{R}^{n^2 \times (1+r)}
\end{align*}
(i.e., also known as the Vandermonde matrix)
and $\mathbf{f} \in \mathbb{R}^{n^2}$ is the vectorization $f\left((UV^\top)_{ij}\right)$.
For the second error, we use Theorem \ref{thm:tensor} to have
\begin{align*}
\mathbf{E}\left[
\left\|
p_r^{\odot}(UV^\top)
- 
\Gamma
\right\|_F^2
\right]
&\leq r \sum_{j=1}^r c_j^2 \ \mathbf{E}\left[
 \left\|(UV^\top)^{\odot j} - T_U^{(j)} {T_V^{(j)}}^\top \right\|_F^2
\right] \\
&\leq 
r 
\sum_{j=1}^r c_j^2 \frac{(2 + 3^k) \left( \sum_{i} (\sum_k U_{ik}^{2})^j \right) \left( \sum_{i} (\sum_k V_{ik}^{2})^j \right)}{m} 
= \| W \mathbf{c}\|_2^2
\end{align*}
where $W \in \mathbb{R}^{(r+1)\times(r+1)}$ is defined as a diagonal matrix with (see Definition \ref{def:not})
\begin{align*}
W_{ii}=
\begin{dcases}
\sqrt{\frac{r(2+3^k)}{m} \left( \sum_{j} (\sum_k U_{jk}^{2})^i \right) \left( \sum_{j} (\sum_k V_{jk}^{2})^i \right)} \quad &i=2,\dots,r+1 \\
0, &i=1.
\end{dcases}
\end{align*}
Putting all together, we have the results, that is, 
\begin{align*}
\mathbf{E}\left[\norm{f^{\odot}(UV^\top) - \Gamma}_{F}^2\right]
&\leq
2 \left( \norm{X \c - \f}_{2}^2 + \norm{W \c}_{2}^2
\right).
\end{align*}
This completes the proof of Lemma \ref{lmm:upperbound}.
\end{proof}

%% file: proof3.tex
\begin{proof}
We denote that 
\begin{align*}
g(\c) := \| X \c - \f \|_2^2 + \| W \c \|_2^2
= \c^\top (X^\top X + W^\top W) \c - 2 \f^\top X \c + 
\f^\top \f 
\end{align*}
for $\c \in \mathbb{R}^{r+1}$ and substituting ${\c^*} = 
(X^\top X + W^\top W)^{-1} X^\top \f$ into the above, 
we have
\begin{align} \label{eq:gbound}
g(\c^*)
&= \f^\top \left( I - X (X^\top X + W^\top W)^{-1} X^\top \right) \f \nonumber 
\\
&\leq \| I - X (X^\top X + W^\top W)^{-1} X^\top \|_2 \| \f \|_2^2 
\end{align}
Since $X^\top X$ is positive semi-definite, it has eigendecomposition as $X^\top X = V \Sigma V^\top$ and $X= \Sigma^{1/2} V^\top$.
Then, we have
\begin{align*}
 X (X^\top X + W^\top W)^{-1} X^\top 
&= 
\Sigma^{1/2} V^\top 
\left( 
V \Sigma V^\top + W^\top W
\right)^{-1} V^\top  \Sigma^{1/2} \\
&=
\left( 
I +  \Sigma^{-1/2} V^\top W^\top W V \Sigma^{-1/2}
\right)^{-1}.
\end{align*}
For simplicity, let $M  :=  \Sigma^{-1/2} V^\top W^\top W V \Sigma^{-1/2}$ 
and observe that $M$ is symmetric and positive semi-defnite because $W$ is a diagonal and  $W_{ii} \geq 0$ for all $i$ 
(see Definition \ref{def:not}).
Then, we have
\begin{align} \label{eq:IminusM}
\norm{I - (I+M)^{-1}}_2 
=
\norm{M(I+M)^{-1}}_2 
=
\frac{\norm{M}_2}{1 + \norm{M}_2} = \left(1 + \norm{M}_2^{-1}\right)^{-1}
\end{align}
since $x/(1+x)$ is a increasing function. From the submultiplicativity of $\norm{\cdot}_2$, we have
\begin{align} \label{eq:M2}
\norm{M}_2
&=
\norm{
\Sigma^{-1/2} V^\top W^\top W V \Sigma^{-1/2}
}_2 \nonumber \\
&\leq 
\norm{
\Sigma^{-1/2} 
}_2
\norm{V^\top}_2
\norm{W^\top W}_2
\norm{V}_2
\norm{
\Sigma^{-1/2} 
}_2 \nonumber \\
&=
\norm{\Sigma^{-1}}_2
\norm{W^\top W}_2.
\end{align}
where the last equality is from $\norm{V}_2 = 1$.
We remind that 
\begin{align*}
W_{ii} = \sqrt{\frac{r(2+3^i)(\sum_j (\sum_{k} U_{jk}^{2})^i)(\sum_j (\sum_{k} V_{jk}^{2})^i)}{m}}
\end{align*}  
for $i \in 2,\dots,r+1$ and $W_{1,1} = 0$. Therefore, 
\begin{align*}
\norm{W^\top W}_2
&=
\max_i W_{ii}^2 \\
&=
\frac{r}{m}
\underbrace{
\max 
\left(
5 \|U\|_F^2 \|V\|_F^2, 
(2+3^r)\left(
\sum_j \left(\sum_{k} U_{jk}^{2}\right)^r
\right)
\left(\sum_j \left(\sum_{k} V_{jk}^{2}\right)^r
\right)
\right)
}_{:= C} = \frac{r C}{m}
\end{align*}
and recall that $\sigma \geq 0$ is denoted by the smallest singular value of $X$.
Then, $\| \Sigma^{-1} \|_2 \leq 1/\sigma^2$.
Substituting the above bounds on $\norm{W^\top W}_2, \norm{\Sigma^{-1}}_2$ into \eqref{eq:M2}, we have 
\begin{align*}
\| M \|_2 \leq \frac{r C}{m\sigma^2 }
\end{align*}
and putting this bound and \eqref{eq:IminusM} into \eqref{eq:gbound}, we have that
\begin{align*}
g(\c^*) \leq 
\left( 1 + \frac{m \sigma^2}{r C}\right)^{-1} \norm{\f}_2^2
= 
\left( 1 + \frac{m \sigma^2 }{r C}\right)^{-1} \norm{ f^{\odot}(UV^\top) }_F^2.
\end{align*}
Combining this with Lemma \ref{lmm:upperbound}, we obtained the proposed bound.
This completes the proof of Theorem \ref{thm:main}.
\end{proof}

%% file: proof4.tex


%
\def\uivj{\u_i^\top\v_j}
Recall that $P:[n] \rightarrow [k]$ is a mapping that satisfies with 
\begin{align*}
&\sum_{i=1}^n \norm{\u_i -\u_{P(i)}}_2 \leq \varepsilon, 
\end{align*}
for some $\varepsilon > 0$ where $\u_i, \v_i$ are the $i$-th row of $U, V$, respectively.
Let $r(x) := f(x) - \sum_{i=0}^r c_i x^i$. Then, 
\begin{align}
\abs{\norm{X \c - \f}_2^2 - \norm{D^{1/2} (\overline{X} \c - {\overline \f})}_2^2}
&= \abs{\sum_{i,j=1}^n r(\uivj)^2 - \sum_{\ell = 1}^k \sum_{j=1}^n |P_{\ell}| \cdot
r(\u_{\ell}^\top \v_{j})^2}\\
&= \abs{\sum_{i,j=1}^n r(\uivj)^2 - r(\u_{P(i)}^\top \v_j)^2} \\
&\leq \sum_{i,j=1}^n \abs{r(\uivj)^2 - r(\u_{P(i)}^\top \v_j)^2} \\
&\leq \sum_{i,j} L \abs{\uivj - \u_{P(i)}^\top \v_j} \\
&\leq L \left( \sum_{i=1}^n \norm{\u_i - \u_{P(i)}}_2\right)\left( \sum_{j=1}^n 
\norm{\v_j}_2\right) \\
&\leq L \varepsilon \left( \sum_{j=1}^n \norm{\v_j}_2\right)
\end{align}
From Lemma \ref{lmm:upperbound} and Theorem \ref{thm:main}, we have that
\begin{align*}
\mathbf{E}\left[\norm{f^{\odot}(UV^\top) - \Gamma}_{F}^2\right] 
&\leq 2 \left( \norm{X \c - \f}_2^2 + \norm{W \c}{2} \right) \\
&\leq 2 \left( \norm{D^{1/2} (\overline{X} \c - {\overline \f})}_2^2 + \norm{W \c}{2} \right)
+ 2 L \varepsilon \left( \sum_{j=1}^n \norm{\v_j}_2\right)\\
& \leq \left( \frac{2}{1 + \frac{m \overline{\sigma}}{r \overline{C}}}\right) \norm{\overline{\f}}_2^2 
+ 2 L \varepsilon \left( \sum_{j=1}^n \norm{\v_j}_2\right).
\end{align*}
where $\Gamma$ is the output of Algorithm \ref{alg:polyts} 
with the approximated coefficient in \eqref{eq:coeffcoreset}, 
$\v_i$ is the $i$-th row of $V$,
$\overline{\sigma}$ is the smallest singular value of $D^{1/2}\overline{X}$ and 
${C}:=\max ( 5 \|{U}\|_F^2 \|V\|_F^2, (2+3^r)  (\sum_j (\sum_{k}{U}_{jk}^{2})^r)(\sum_j (\sum_{k} V_{jk}^{2})^r))$.
This complete the proof of Theorem \ref{thm:coreset}.